**"In folly ripe. In reason rotten"[1] Putting machine theology to rest**

Mihai Nadin, Institute for Research in Anticipatory Systems, University of Texas at Dallas, Richardson TX USA

**Abstract**: Computation has changed the world more than any previous expressions of knowledge. In its particular algorithmic embodiment, it offers a perspective, within which the digital computer (one of many possible) exercises a role reminiscent of theology. Since it is closed to meaning, algorithmic digital computation can at most mimic the creative aspects of life. AI, in the perspective of time, proved to be less an acronym for artificial intelligence and more of automating tasks associated with intelligence. The entire development led to the hypostatized role of the machine: outputting nothing else but reality, including that of the humanity that made the machine happen. The convergence machine called *deep learning* is only the latest form through which the deterministic theology of the machine claims more than what extremely effective data processing actually is. A new understanding of complexity, as well as the need to distinguish between the reactive nature of the artificial and the anticipatory nature of the living are suggested as practical responses to the challenges posed by machine theology.

---

[1] Raleigh, Walter (1596)

"…it's just a block of wood! I burned half of it for
heat and used it to bake my bread and roast my meat.
How can the rest of it be a god? Should I bow down to
worship a piece of wood?"[2]

# 1 Introduction/Preliminaries

A distinguished colleague (holding an endowed chair at an Ivy League university), known for his work in computational molecular biology, rushed a kind note to me: "Your anticipatory research theme is now a thematic focus in IARPA program: Anticipatory Intelligence. The world is catching up with you." On the IARPA website I read: "Anticipatory intelligence focuses on characterizing and reducing uncertainty by providing decision makers with timely and accurate forecasts of significant global events." As well-intended as the congratulation was, it brought up many instances of no less well-intended but deceiving use of the word "anticipation." "Have you anticipated this?" would be the well-meant comical remark that gives away the fact that the person asking had no idea what anticipation is. Few really do. IARPA certainly does not. Prior to IARPA—and to the seductive task of reducing uncertainty through "anticipation"—*intelligence* had the same privilege, but under DARPA.

This episode conjures questions such as: "How do scientific concepts make it into everyday language?" "How does a certain target—intelligent behavior—morph into whatever it can be confused with?" And more important, "How consequential is science anyway?" Albeit, if what scientists try to accomplish, sometimes under misguided assumptions, is not consequential in some ways (including why something might not be achievable, now or in the foreseeable future), the whole enterprise ends up as an exercise in futility—for which society pays in one way or another.

---

[2]From the Book of Isaiah, the Prophet, Chapter 44:19-20 (abridged).

If nothing else, the quest for knowledge was always associated with rationality. Still, a short time ago, a rather successful innovator (Elon Musk 2014) went as far as to state that, "With artificial intelligence, we are summoning the demon." (Consequential to the extreme, some would say). "In all those stories where the guy with the pentagram and the holy water, it's like 'Yeah, he's sure he can control the demon.' Didn't work out." Stephen Hawking joined in the expression of concern. Ray Kurzweil read something else into it (presumably also consequential): "We will be able to upload copies of our brains to intelligent machines and thus achieve digital immortality," (Kurzweil 2005, 2013)[3]. One of the leading technologists in the autonomous vehicle (i.e., self-driving car) "mini-revolution," Anthony Lewandowski (a member of the Google and Uber "nobility") formed Way of the Future, registering it as a tax-exempt religious organization: "…through understanding and worship of the godhead, contribute to the betterment of society" (Harris 2017). As divergent as such positions are, they ultimately tie into what this study defines as the theology of the machine, ubiquitous not only in the world of computation, but also within science in general. To return to the subject that has triggered so many controversies—the Dartmouth Conference of 1957—artificial intelligence proved to be consequential beyond the hopes of those who initiated it, though not necessarily in the way they wished. AI, for short, even became a new 2-letter word since that time. Parallel to the Dartmouth event, *Desk Set*, a film, directed by Walter Lang, featured Katherine Hepburn as Bunny Watson, the "living" prototype of the computer that could answer any question, and which could have replaced her as reference librarian at the Federal Broadcasting Network. The real Watson, claiming AI capabilities, won the *Jeopardy* contest (January 2011) with living competitors who were famous as the most successful contestants (Ken Jennings and Brad Rutter). The Watson of IBM (which supported production of the movie) is by now available for hire ($265 per month) to offer "a cognitive computational self-service experience that can provide answers and take action." Celebrated as "the most well-known example of artificial intelligence in use today," it recently came under scrutiny. Cory Doctorow (Boing Boing, November 13, 2017) writes: "Watson for Oncology isn't an AI that fights cancer, it's an improved mechanical Turk that represents the guesses of a small group of doctors." But on the heels of this comes CheXNet: Radiologist-Level Pneumonia Detection on Chest X-Rays with Deep Learning (https://stanfordmlgroup.github.io/projects/chexnet/) claiming better performance than radiologists. It is worth checking on claims because they usually divulge premises upon which the "miraculous" performance is (or not) achieved..

---

[3] Inspired by this notion, Tod Machover, of the MIT Media Lab, wrote and composed the opera *Death and the Powers* in 2011.

## 2 Little Money for a Very Ambitious Project

John McCarthy called it "artificial intelligence." The participants at the 1956 Dartmouth Conference (formally, the Dartmouth Summer Research Project on Artificial Intelligence) agreed: "…every aspect of learning or any other feature of intelligence can be so precisely described that a machine can be made to simulate it." Machine meant "an agent that manipulates symbols." The extended workshop at Dartmouth College (where McCarthy, who applied for funds at the Rockefeller Foundation, was teaching) was documented in detail by Ray Solomonoff (1956). "This new field of mathematical models" (as the Rockefeller Foundation understood it) was "difficult to grasp" and only half of the rather modest funding request was approved with "a great deal of hesitancy." We even know that at some moment the future celebrities of AI checked for the meaning of the word "heuristic" in a dictionary, wondered about Ashby's description of the *homeostat*, considered (as as subject of interest) machines playing chess, and also programming languages for more ambitious goals. It seems that those philosophically oriented (Simon, Newell, as well as McCarthy and Minsky) agreed that artificial intelligence could save philosophy from insignificance. Indeed, many philosophic subjects, such as thinking, in particular deductive and inductive reasoning, but also the more recent probabilistic reasoning, permeated the agenda. But more significant is the deceptive echo of Hilbert's challenge (the famous *Entscheidungsproblem*) that eventually led to the Turing machine—the actual star, or better yet, god, of the entire event. Mathematics TM—for Turing Machine—was also a subject. Minsky came up with a "geometry machine" able to prove theorems; Newell, Shaw, and Simon, with a "Logic theorist" (forerunner of the General Problem Solver). Warren McCulloch claimed, without anybody questioning him, that the human brain is a Turing machine.

The subsequent history of AI is pretty well documented (Lungarella**,** Iida, Bongard**,** Pfeifer 2007): the victory of computation, with all its desired and (most of the time ignored) undesired consequences. Computation became the underlying foundation of a new civilization (Nadin, 1997). Of course, the prophets of the movement—those self-declared, as well as those who earned their recognition—are celebrated. Day-in-day-out we learn of the revelations associated with Big Data processing, as well as of the deep, deeper, and ever more deep learning that defeats champions (of checkers, chess, Go, e-Sport). It out-diagnoses the medical profession and produces more impressive art than artists do (according to a report entitled "Computers can now paint like van Gogh and Picasso," Murphy 2015). Move over Rembrandt, Matisse, Picasso, Pollock and the rest, the Convolutional Neural Network is coming (Gatys, Ecker, Bethge 2015).

Neural networks "learn learning" (Andrychowicz et al 2016) and design new networks (Zoph and Le 2016). The Rapture is imminent. It is the heaven of singularity (Kurzweil 2005) when no distinction between human thinking and machine intelligence will be possible. Better yet: *When Will AI Exceed Human Performance? Evidence from AI Experts* (Grace, et al 2017). No less than 352 researchers (from the 1634 contacted) were questioned on the probability of high-level machine intelligence (HLMI), settling in the years to come. HLMI "is achieved when unaided machines can accomplish every task better and more cheaply than human workers" (Grace et al 2017). Since science is considered an endeavor of affirmation, even the suspicion of doubt, not to say negation—"It's not possible …" (you fill in here the blank with whatever someone might assume computation cannot accomplish, let's say to have sex, to give birth, or to establish peace on Earth)—could become anathema. *Dubito ergo sum* was okay for Descartes, but you'd better stay away from the hall of mirrors (digital, of course) if you question whatever the dogma of the day is.

Time to take a deep breath: Isn't today's Turing machine epiphany (this word is not chosen by accident) the outcome of impressive proofs that demonstrated the impossibility—which is a negation, i.e., well more than a doubt—of a mechanical procedure for determining the truth values of mathematical statements? The outcome of this particular challenge:

1) Church, based on papers by Gödel, showed that "the quest for a general solution of the decision problem must be regarded as hopeless" (Church 1936 a, b, c);

2) Turing (1936-7) proved that the Hilbertian *Entscheidungsproblem* has no solution;

3) Gödel documented that Hilbert's goal[4] cannot be reached, i.e., that consistency and completeness of some formal systems is unattainable (Gödel 1936). To this distinction, pertinent to formal systems in Gödel's proof, we shall return, reaffirming its validity to distinctions in reality (Nadin 2014).

The negation in reference to the mechanical decision procedure is indirectly an affirmation of what became the algorithmic set of rules making the solving of a problem through mechanical "reasoning" possible. Indeed, there is a part of reality that can be described through algorithmic computation. It turns out that this part of reality is at the same time decidable: it can be fully and consistently described (Nadin 2014). Those who pay attention to the details of Hilbert's challenge understand that *Verfahren* (procedure) is not really the same as *algorithm*. "Who from among all of us would not gladly lift the veil under which the future hides?" was Hilbert's rhetorical entry to presenting some challenges to the community of mathematicians. Little could he know that those (Church, Turing, Gödel) demonstrating the impossibility of a machine-based procedure for proving the truth of mathematical statements actually set the foundations of a particular type of machines that will eventually change civilization. *Homo Turing*, as I would call him or her, is an outcome defined by Bolter (1984): utilitarian, calculating, shallow, living by cost-benefit analysis. It seems that in reshaping *homo sapiens* intuition, spontaneity, empathy, compassion, and even judgment were traded for expediency. Of course, Turing could not foresee the consequences of his visionary work.

---

[4] Das Entscheidungsproblem is gelöst, wenn man ein Verfahren kennt, das bei einem vorgelegten logischen Ausdruck durch endlich viele Operationen die Entscheidung über die Allgemeingultigkeit Erfüllbarkeit erlaubt, (Hilbert, Ackermann 1928).

This outcome, probably more relevant to an anthropological account, is related to the foundational work we are examining. Among the many consequences of this foundational work, two are of immediate significance:

1) The construct computation and the associated domain of the computable was established as a distinct epistemological domain. It has a rather long history (going back to pebbles, knots, beads on the abacus, mechanical devices, etc.), being, in the final analysis, one among many representations on whose basis knowledge acquisition, dissemination, and evaluation take place.

2) In contra-distinction to the computable (identified, without any reason given, with the Turing machine, i.e., algorithmic computation), there is the non-computable (at least non-algorithmic), to which the Entscheidungsproblem, i.e., a mechanical (that is, automated) procedure for determining the truth of mathematical statements, belongs. The subdomain of algorithmic computation became the placeholder for all forms of discrete computation, and the underlying computation of what is defined (or maybe not at all defined) as AI.

If nothing else can be derived from these accepted discoveries, one statement stands out: human beings, in their quest for understanding the world, constitute themselves through their activity, testimony to their abilities. Alas, they prove theorems, but not in a mechanical (i.e., machine-based) manner. Moreover, they are not subject to the infinite loop of the halting problem: that is, can a computer recognize when the programs task is finished (or will it continue to process indefinitely)? The human being, and for that matter any form of life, independent of the activity through which it expresses itself, would halt. In other words, it *understands* whatever is performed and stops, either when it cannot achieve what it wants or after achieving it. Based on these two observations, one can infer that, contrary to statements made since Dartmouth, human beings are not reducible to algorithmic machines. That the situation requires a more nuanced approach will become evident.

## 3 The Heresy of Questioning TM Reductionism

Even those who conceived the new machines were surprised, in the various phases of the ascertainment of algorithmic computation, by their performance. This is understandable. Wonders are phenomena for which we are not prepared, neither in terms of our ability to understand our own actions and ideas, nor in terms of our emotional reactions. But there is catching up, and what seemed out of proportion—algorithmic computation by now has a history of extraordinary achievements—is slowly integrated in culture. The cell phone brought computation to everyone's pocket. The ubiquity of algorithmic computation rivals that of electricity (the almost prophetic view of Mark Weiser 1991). Something else is happening as well: "mistaking the abstract for the concrete" (Whitehead 1992, p. 2), also known as *to hypostatize* or *reification*, to construe something. The classic example associated with this form of misrepresentation is that of Hegel: the real world is the creation of the idea. (Debunking Hegel, Marx hypostatized the material world.) And before that, religion promoted the godly view of the world. i.e., construed it as created by some divinity. Isaiah's description in the motto to this study says it all: we made gods and ascribed our own rules to them. For those following in the footsteps of religion, or of Hegel's idealism, or of Marx's materialism, the abstract Turing machine (the idea, as Hegel would call it) also creates a reality. The abstract mechanical procedure for ascertaining the truth of statements other than mathematical is mistaken for the concrete, for the reality of how humans think. The theoretic representation of this reasoning process leads to a religion of no distinction between machines (instantiating the abstract mechanical model) and human beings. This is how deities emerge as an explanation for phenomena otherwise impossible to explain at a certain moment in time. The consequence is relatively straightforward: those who question the reduction of thinking to the functioning of the algorithmic machine, i.e., those who undermine the TM idolatry, are seen as heretics.

In the spirit of placing the memorable Dartmouth moment, when symbolic processing emerges as the assumed embodiment of artificial intelligence, in the broader context of algorithmic computation, it is time to see what happens to the various heretics who challenged the new religion of the machine and AI. Of course, Hubert Dreyfus and Joseph Weizenbaum come first to mind, as do the high priests of an intolerant intellectual inquisition whose activity transcended the usual academic infighting. Reading today Dreyfus's *Alchemy and Artificial Intelligence*—the RAND paper P3244 (Dreyfus 1965)—and Papert's (1968) *The Artificial Intelligence of Hubert L. Dreyfus. A Budget of Fallacies* (Artificial Intelligence Memo No. 154) suggests a very good idea for a movie in the spirit of *Inherit the Wind* (focused on the introduction of Darwin's evolution theory in America education). The same holds true for Weizenbaum's *Computer Power and Human Reason* (1975) and McCarthy's "An Unreasonable Book" (1976). There is no reason to demonize and no reason to idealize anyone. By now, most of the community of computer scientists set aside the arguments and counterarguments of those confrontations. Just as a reminder: Hubert Dreyfus, respectable philosopher and admirable teacher, ascertained that The analogy brain-computer hardware and mind-computer software is a misleading assumption; the same holds true for the assumed discrete computation driven by algorithms on symbolic representations (1972).

Behind these premises, which he discussed in detail, although somehow imprecisely, are the far-reaching views according to which the dynamics of reality can be described through predictive rules or laws. The expectation of prediction is the outcome of the doctrine that reality can be reduced to its physics. We shall return to the focus on physical symbol processing, as AI processing was construed. It is the Achilles heel of the arguments advanced by those who promoted symbol processing as the foundation of AI (in particular, Newell and Simon), but also of the views advanced by Dreyfus and others who disputed the context-free processing of data as a path towards emulating intelligence.

This summary does not do justice to the many distinctions that Dreyfus advanced, and even less to the richness (depth and breadth) of the arguments. Of course, it spares Dreyfus the humiliation of quite a bit of hasty philosophizing, sometimes even to the detriment of those whom he quoted or alluded to (in particular Heidegger and Merleau-Ponty).

Weizenbaum knew a lot about computers—from the analog to the digital, including neural networks—but not as much about philosophy (although political science was among his interests). Israel Scheffler (2004) presents a good portrait of Weizenbaum's philosophy, not to be downplayed by those who saw in him only the MIT professor of computer science. He did not exclude the possibility of AI, but claimed that with larger and larger programs, more and more entangled, it becomes very difficult (if not impossible) to distinguish between desired outcomes and possible malfunctioning of extreme consequences. Weizenbaum specifically associated decision-making with computation, but argued that choice is a human capability, not within the possibilities of digital processing. Computers have no wisdom or compassion, which, in his view (passionately ascertained), are part of human intelligence. The same applies to the role of emotions. There are dangers to be aware of when the programmed society replaces the society of informed decision-making.

Were it only for the sake of rehashing what was said and written at that time, the most we could expect would be a contribution to the culture of the civilization of computation. Not unimportant, but of no consequence for what defines current views of reality, of the human being, of science and technology. For suggesting that computers, in whichever form, could not do something—whatever that might be—Dreyfus and Weizenbaum and many others were treated like intellectual Luddites. For ascertaining that the brain is not a computer and intelligence is more than solving problems based on rules, such authors were ridiculed by colleagues enjoying positions of authority. The fact that none of them realized what Turing, Church, and especially Gödel established—that there are tasks for which the algorithmic, at least in its current expression, is not adequate—is an observation impossible to ignore. More important: in the spirit in which determinism was hypostatized and became the religion of science, some of Dreyfus's and Weizenbaum's colleagues effectively promoted the theology of the Turing machine. The various commandments attributed to the divinity by the humans who constructed it (in search of answers to questions for which no better answers could be given at the time) make up what is known as religion. Churches promote such commandments as divine. In a strange parallel development, Descartes's views, proclaimed but never proven, gave rise to the theology of reductionism and determinism embodied in the machine, extended to one particular type: the Turing machine. There is historic precedent to this: the hypostatized brain as made of clay infused with spirit; the brain as a hydraulic machine (the humours, fluids running though the machine);

the mechanical automata model (more like a clock); Hobbes's mechanical motion brain;

Helmholtz's "neural network," pretty much like the telegraph; the brain as a quantum computer

(Fig. 1[5]). It never ends (Zarkadakis 2015). The obsession with reducing the brain to the machine-

*du-jour* is as understandable as it is infantile.

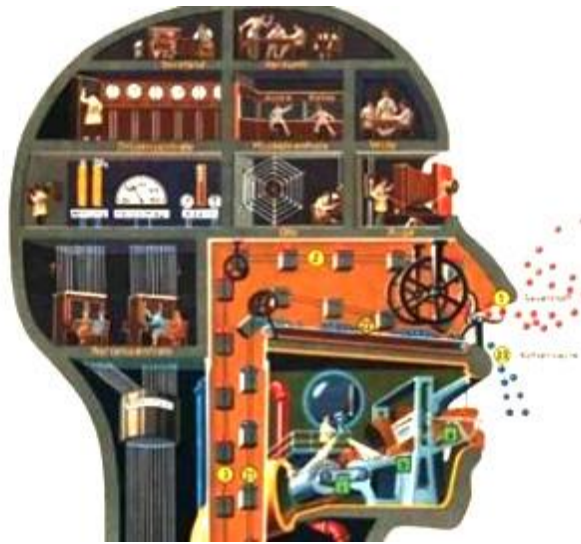

**Figure 1** Representations of the brain and human as machine (Fritz Kahn, Body Machines[5])

These are steps on a long journey, never to come to an end, since, as we explain ourselves, we

change. In this process, algorithmic computation is only an intermediate phase, as *S*is also only

one step in an open-ended sequence of the making and remaking of the human being. To the

science to come, looking back at the times when AI was a new topic (even the sponsors of the

Dartmouth Conference regarded it with some suspicion, as we have seen), will bring smiles pretty much like ours when we read about clay and the spirit, the humors, the hydraulic pump model, etc.

---

[5] Fritz Kahn. Der Mensch als Industrie Palast (Man as Industrial Palace), 1927

What today intrigues a reader of the words said and printed at the time Dreyfus confronted the AI experiment is the "proof by intimidation"—an expression that McCarthy brought up, complaining that Weizenbaum used some words as clubs. Dreyfus knew close to nothing about computers, and Weizenbaum did not produce another ELIZA (which is still successful in our days) as he was contrasting human reason to the power of computers. But their arguments were about principles—many insecure characters feel intimidated by principles. To discuss Dreyfus as a chess player, only because he discussed computer-based chess playing, is to miss the point he was making. It is unfair to generalize from Dreyfus losing to a machine, or even from the ex-world chess champion Kasparov losing to Big Blue. They were playing against hundreds, if not thousands of adversaries who cooperated in investigating the huge space of possibilities associated with the game. If Dreyfus, and later Kasparov, had had access to all the resources that the chess program was using—the knowledge database of openings played by 100 grandmasters, fast search, pattern recognition, etc.—they, or anyone else, would have been competing in a fair contest. The human-hours of development that went into having a brute force program win a chess game exceeds not only the lifespan of one player, but probably that of all players. (Kasparov learned that the machine he was playing against was assisted not by one, but by two grand masters.) Be that as it may, McCarthy argued for better logic, not for more powerful engines when he attacked Weizenbaum's warning about a time when machine brute force will take over human intelligence. Worse yet: the machine's winning and intelligence are incongruent.

Let's examine Papert's argument against Dreyfus. Inspired by Dreyfus, Huston Smith (a professor at MIT sympathetic to Dreyfus) submitted a typescript to Papert. He mentions, "It has been estimated that there are $10^{120}$ different paths through a complete chess maze." This is, of course, the Shannon (1950) number: the lower-bound of the game tree. It is based on an average of ten thousand choices. Time for Papert's irony (not realizing that Shannon was behind the calculation): "Prof. Smith quite naturally relied on his colleague Dreyfus, who is reputed to be an expert on computers." The final proof is in the numbers: if five moves (on average) can be made and "one can explore all possibilities to a depth of as many as twenty moves" (Dreyfus 1965), it would take 32 trillion microseconds (i.e., a year) for the computer to analyze a game situation. Papert, the mathematician, was either wrong or disingenuous in the argument. Looking ahead in a game of chess is a matter of permutations. Therefore, the five moves, as an average, that can be made while looking ahead twenty moves would indeed mean $6^2 35^5$ or even, to quote from Papert, $2^5 2^{20} 2^{20}$. Had Papert taken advice from one of the freshmen ("with some facility in arithmetic and elementary knowledge of computers")—as he ironically formulates a suggestion to Huston Smith—he would have noticed that with each move, as the game advances, the numbers change very fast. Actually, the game has a powerful convergence: even the most disputed games do not exceed 40-41 moves. The chess games database of 685,801 games calculates 40.04 as the average number of moves.[6] It is the case that between chess players of comparable skill, the one playing the white pieces wins most of the time. (The first serve in tennis is the easiest analogy for games played more and more "mechanically".)

The point Papert was making totally missed Dreyfus's argument: a checkers or chess player operates in a field of meaningful searches. The machine, as a purely syntactic device, does not make meaningful decisions; it is processing data with the aim of reaching a well-defined target. Brute force associated with improved search functions (improved heuristics) is deployed for calculating, but not for "playing chess." The program activates the hardware, not unlike the fingers of an operator activates an abacus. Understanding what each step means is neither possible nor really attempted. To use a figure of speech, they are practicing the Entscheidungsproblem on a decidable mathematical problem (the chess game). Chess, as a cultural artifact, is not a math problem. This realization is of no consequence in writing the software and designing the search facilities. It all becomes a matter of pattern recognition, for which intelligence is required. A game with its many dimensions—cognitive, aesthetic, emotional, etc.—engages the person on many levels. The desire to understand—which is a prerequisite of intelligence—harkens back to what prompted the invention of the game in the first place. The game's competitive nature translates into engaging the player as a whole. As IBM reported, in the re-match, Kasparov simply didn't feel like playing. Deep Blue does not have (and does not need) a "feel"-like utility, or, for that matter, nothing, except for scoring, that might testify to what it means to be successful. Automating the playing of a game that involves the intelligence of living competitors is not the same as making intelligence available as a utility (or commodity). Intelligence is always about meaning.

---

[6] www.chessgames.com

On this note, Papert and Dreyfus are in the same boat: neither is aware of, or even marginally interested in, semiotics. As a matter of fact, all those advancing the idea of artificial intelligence as outcome of symbolic processing missed the Charles Sanders Peirce moment in the history of science and philosophy in the USA. Moreover, the pioneers in computers were not better. This is surprising since Arthur W. Burks, for example, who helped in building ENIAC, the first general purpose digital computer, was more than aware of Peirce.

**4 Stating vs. Proving**

Allen Newell and Herbert A Simon stated, "A physical symbol system has the necessary and sufficient means for general intelligent action" (1976). This is known as the physical symbol system principle. It was never proven. Some of the terms are questionable, such as physical symbol, or general intelligent action. The concept of *system* is used in quite a vague manner. What kind of system? Intriguing also is the psychological assumption: "This is how individuals process symbols." There was no proof for that at that time—as there is none today. Psychology was not known for its interest in semiotics, and actually discarded it (Bell 2005). Given the rather questionable condition of psychology—in a very charitable description, a discipline oscillating between the practical (psychological advice to patients or institutions) and the meta-theoretical (it's neither physiology nor brain science, but theoretizes their content matter)—it is surprising that science has given it more than considerate attention, and let it become the monster it is today. It forced upon society the IQ, which contaminated to a large extent the research of intelligence (for which it is supposed to stand). Between Freud's hoax, the IQ (enlisting statistics, of course) and the so-called prospective psychology (in which the Templeton Foundation found a flag bearer), there is little, if any, difference. The problem is that machine theology often invokes the circularity of psychological reasoning: "This is how humans do, so….", if the word reasoning can be connected to it.

Turing himself used the word *symbol* in describing his machine: "an unlimited memory capacity obtained in the form of an infinite tape marked out into squares, on each of which a symbol could be printed" (Turing 1948, p. 3). The only requirement is that the symbol come from a finite alphabet.

The formal definition (Hopcroft and Ullman 1979, p. 148) of a one-tape machine **M** is a 7-tuple:

$$M = <Q, \Gamma, B, \Sigma, \delta. q_0, F >$$

in which

$Q$        finite, non-empty set of *states*;

$\Gamma$        finite, non-empty set of *tape alphabet symbols*;

$b \in \Gamma$     *blank symbol* (often allowed to occur infinitely);

$\Sigma \subseteq \Gamma \backslash \{b\}$ *input symbols*, allowed to appear in the initial tape contents;

$\delta : (Q \backslash F) \times \Gamma \rightarrow Q \times \Gamma \times \{L, R\}$    *transition function* in which L is left shift, R is right

       shift.

$q_0 \in Q$   *initial state*;

$F \subseteq Q$     set of final states or accepting states. The initial tape contents is said to

       be accepted by M if it eventually halts in a state from F.

While all the terms that Turing used, or all the concepts in the 7-tuple formal definition are well defined, the notion of the symbol is not.

4.1 Symbolic representation

For all we know, the word *symbol* was used quite a bit in a variety of texts, from those of religious intent to the more ambitious attempt to define *culture as expressed in symbols*. It is doubtful that etymology (the classic reference to ancient Greek, for example) would shed light on the matter. Words mean what we want them to mean, or what the context of their use suggests;

for example, the meaning of *quantum leap*, or even the meaning of computation. Many steps further down the epistemological road from Turing came the already-mentioned assertion, grounded in psychology, that processing physical symbols is the foundation of intelligent-like performance by something other than the human being. Candidly speaking, those who focused on processing physical symbols were considering what it would take to achieve artificial intelligence, not what kind of machine would eventually accomplish it. Dreyfus's reaction to Newell and Simon—the Rand Corporation moment (where they developed some of their ideas and Dreyfus disputed them)—was not about computers, but about the conceptual premise: "…since Descartes…understanding consists in forming and using appropriate symbolic representations," (Dreyfus and Dreyfus 1986). Of course, this goes back even farther in time, to Plato's views, where "rules of functioning in the expert's mind whether he is conscious of them or not." (But this is not the place to rewrite the history of rationalism.) However, in their opposition, neither Newell, Simon, and McCarthy et al, nor Dreyfus and his many supporters take the time to define *symbol* or *symbolic*. C.S. Peirce would have been the most pertinent reference. Although his work was marginally acknowledged, he was not unknown. As I already mentioned Burks was knowledgeable in computers and Peirce. Before all of them, Peirce was involved, as were, many years earlier, Pascal and Leibniz, with "Logical Machines." In a short article (1887), Peirce makes reference to *Gulliver's Voyage to Laputa*.

> In the "Voyage to Laputa" there is a description of a machine for evolving science automatically. "By this contrivance, the most ignorant person, at a reasonable charge, and with little bodily labor, might write books in philosophy, poetry, politics, laws, mathematics, and theology, without the least assistance from genius or study."

For his thoughts on the matter, Peirce was even eventually credited (Dalakov, History of Computers website) for having "invented in 1886, together with Allan Marquand (his student at Johns Hopkins University) "the first electrical logical machine." Be this as it may, what should have been acknowledged is Peirce's semiotics ("logic of vagueness" was his actual target, Nadin 1983).

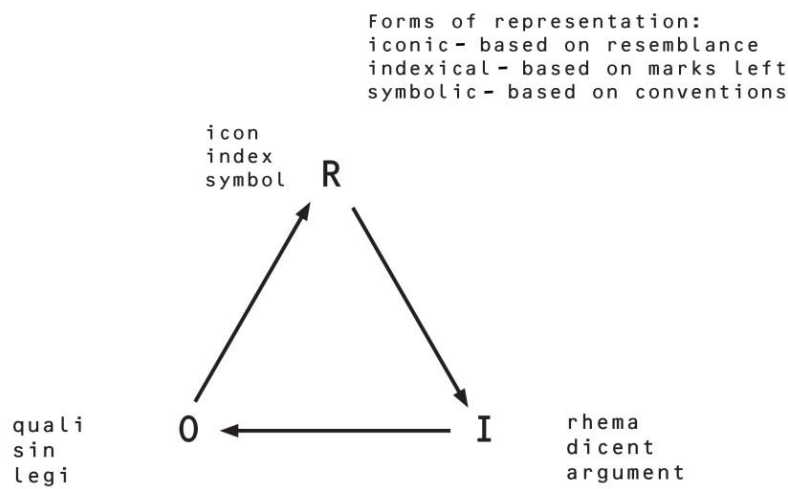

**Figure 2** A coherent theory of semiotics within which symbolic representation is well defined

To argue in favor of or against Plato's theory of forms, or, for that matter, in favor of or against Leibniz (Dreyfus called him the grandfather of expert systems), to argue with the participants at the Dartmouth conference is really fighting windmills, since their most important concept—the symbolic—is ill defined. With this in mind, the Dreyfus-Weizenbaum critical position in regard to machines and intelligence deserves to be reconsidered. Within the "winner-take-all" model, they are seen as sad cases of missing the train that brought AI into the forefront of computation. Only fools argue with success. But is it success?

In Peirce's view, the symbolic is nothing more than a form of representation. To be more precise, the overreaching concept is that of a sign:

I define a sign as anything which is so determined by something else, called its object, and so determines an effect upon a person, which effect I call its Interpretant, that the latter is thereby mediatily determined by the former (EP2, 478, 1998)

In diagrammatic expression (Fig. 2), the sign is the *unity* between what is represented (the Object, a placeholder for anything in the world, things, processes, ideas, etc.) something that stands for the object, called representamen, and the open-ended process of interpretation. Of course, each interpretation becomes yet another Object, for which *representamina* will stand, and so forth. Within Peirce's triadic-trichotomic structure, Objects can be represented in their quality, uniqueness, or necessary (law-like) nature. The representation can be iconic (based on resemblance), indexical (mark left, such as a fingerprint), or symbolic (by agreement, convention). Interpretations can result in rhematic expression (the "aha" moment), dicent forms (ascertainments), or arguments (such as logical deductions). The nature of the sign process is such that at this level, only a limited number (10) of possible signs is possible.

This is not a lesson in Peircean semiotics (which would require a broader context, more precisely placing it in the larger framework of his philosophy). Rather it is an attempt to give the notion of *symbolic* some underpinning. In the absence of this effort, the affirmation of AI as physical symbol processing resembles the dialogs in the theater of the absurd (Ionesco, Becket, etc.): characters talk past each other; each has something else in mind, although what they say resembles the common use of language. Peirce placed the sign at the center of his semiotics, which is meant to transcend the exclusive focus on words usually understood as symbols. He wrote about the inclusive nature of signs. More precisely, symbolic representations include iconic and indexical aspects. The words include assertions and judgment, and thus facilitate the speech act through which knowledge is formulated and shared.

The focus of this part of the broader discussion on the theology of the machine is on the notion of symbolic processing, more precisely, its meaning. To anchor the argument in a clear conceptual framework is not optional. Peirce's semiotics, ignored or not, is a necessary reference. Two aspects beg our attention:

a)   How does semiotics, the discipline of sign representation, influence conceptions of the nature of computation and the nature of intelligence?

b)   How does semiotics inform the scientific discussion of the possibility to replicate intelligence in some medium other than living matter?

4.2 A short detour

Before addressing each, let's return to the conceptual conflict described.

I take the liberty of quoting from conversations with distinguished colleagues (each deserving respect and admiration, regardless of whether we agree or not with what they say):

*I recently had a chance to refresh my memories of Hubert Dreyfus, and step back to see what kind of real influence he had. The answer is, he was inconsequential. He posed no problems that AI did not, in its own way, answer, and he missed the really big questions.*

*If you think of any philosophers as having "influence," John Searle's Chinese Room puzzle raised interesting questions about understanding (though he failed*

*to understand what understanding is himself) and it was funny when that too was put to rest. The only philosopher people in AI really paid attention to was Daniel Dennett, who raised interesting questions knowing that they might not be answered to everyone's satisfaction anytime soon. He thought—he thinks—it was brave of AI even to take on such difficult issues, since most philosophers were still enchanted with their own hand-waving assertions. (Pamela McCorduck, e-mail of September 8, 2017)*

*Dreyfus was the primary interpreter of Heidegger from within an American cognitive framework. He worked to recast Heidegger's insights into terms that would be comprehensible by people with a scientific /engineering background. This is what Flores and I picked up on, focusing on this analysis of thrownness and the contrast of ready-to-hand and present-at-hand, as a way to orient our approach to computer systems*

*I know there are many other aspects to Heidegger's work that did not find their way into our picture, and maybe only partially into Dreyfus' picture. (T. Winograd, e-mail of August 3, 2017)*

There is a large body of commentaries on the subject, and there is a lot to acknowledge in respect to the Weizenbaum-Dreyfus moment as it pertains to AI and computation in general. The reason to remember is not to adjudicate victory for somebody, or even something—AI is most spectacular in those days, and Big Data as a subject of computation (and huge contributor to work in neural networks) once again open new perspectives.

The reason is the need to understand, which means to intelligently assess not only successes, but also perspectives. Semiotics and Peirce came up because way back, at the Turing moment, the symbol was called into the existence, as a mark on a square on the infinite tape, in relation to the effective procedure, i.e., algorithm embodied in his machine. What happened since, in terms of machine performance, cannot be admired enough, while at the same time, what did not happen—because of the lack of a broader view—cannot be ignored, since quite a bit of the knowledge that would have been necessary for even more impressive progress was and is available. That knowledge would also help in avoiding the rather disturbing consequences of computational fanaticism.

## 4.3   Significance

In bringing up semiotics, we suggested only the least controversial aspect. Indeed, not unlike all the constructs deployed in order to capture quantitative aspects of reality, such as numbers, signs are as well the outcome of epistemological activity—i.e., how we get to know what we want to know. There was semiotics before Peirce, and more of it was stimulated by his views. What comes into focus with semiotics is the relevance of meaning, complementing that of quantitative descriptions. Moreover, the meaning of numbers—a construct based on symbolic representation—associated with measuring quantitative aspects of reality, becomes accessible. So does the meaning of words in what is called *natural language*, of what people see or hear, but also of images and sounds, of models and simulations, of everything constructed by the living in order to represent the environment of its existence.

Neither Plato, nor Leibniz, searching for a universal language and even seeking an "algorithm" for proving logical statements, and even less McCarthy, Minsky, Papert, Newell,

Simon, et al. were wrong in assuming that processing a certain form of representation—i.e., the symbolic—plays an important role in intelligent expression. The understanding of reality is mediated through shared representations. Rather, they promoted an incomplete understanding of the role of representation in knowledge acquisition as well as in knowledge expression. That the conversation between those ascertaining the role of symbolic processing and those questioning it took a nasty tone is indicative of the fact that there is a lot of insecurity at the limits (as relative as they are) of our knowledge. To give one example: Is intelligence contingent upon embodiment—the discussion of the intelligence characteristic of ice skating, driving, swimming, etc.—or independent of body expression? The early AI proponents ascertained an understanding of intelligence as only the outcome of processing symbols. Only later did Rodney Brooks confirm Dreyfus. Evidently, Humberto Maturana, Francisco Varela, Eleanor Rosch, and so many others had no difficulty in accepting the rationality of embodied intelligence.

With Peirce in mind—i.e., integrating the semiotic perspective—the subject is reframed: Representations of all kinds come together, or are checked one against the other. The word *ice-skating*, or *biking*, is insufficient for performing the action, which involves not one, but many expressions of explicit and implicit knowledge. In other words, a variety of types of intelligence, not the generality of intelligence set as a theoretical and practical target by the initiators of the AI program have to be acknowledged. They all reflect anticipatory actions pertinent to self-preservation of life.

**5 Machine Reductionism**

On this note, the discussions of yesteryear take on new meaning, and the major issue is again with us: machine reductionism. Given the spectacular performance of AI in its connectionist embodiment—machine learning, in particular deep learning—it is probably less dangerous, in respect to maintaining clarity, to proceed step by step. Instead of focusing on the new "records"—chess or Go, image identification, or "learning to learn," etc.—we can benefit from reconsidering an accomplishment rightly celebrated (with a Nobel Prize in 1963) that has since inspired the scientific community. This is machine reductionism at its best. The Hodgkin-Huxley mathematical model (1952) suggests a physics perspective of the physiological process of initiating and propagating action potentials in neurons. The experimental subject was the Giant Axon of Coligo. Its large diameter offered the advantage of affording measurements of electric variables. As the diagram shows (Fig. 3), the membrane can be modeled as an electrical circuit.

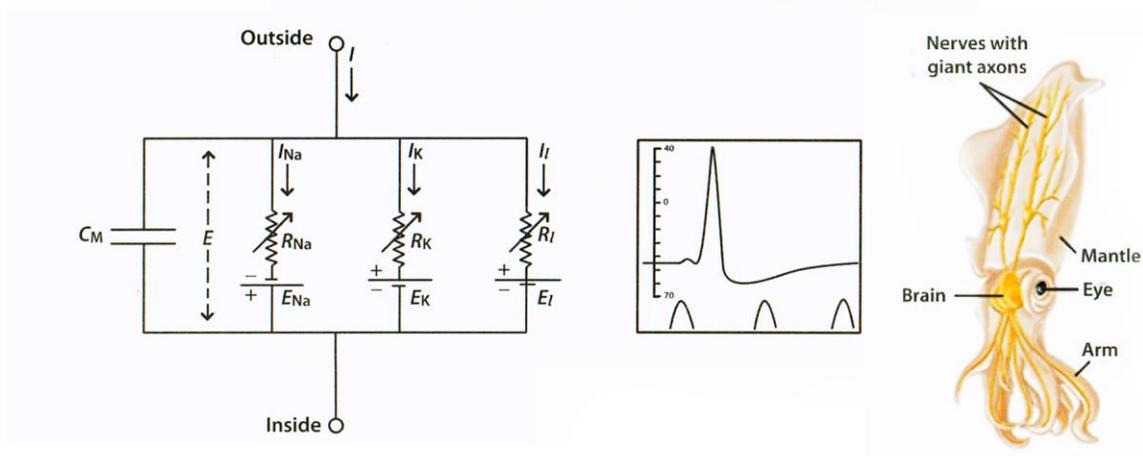

**Figure 3** The limits of reductionism—a Nobel Prize example. Curve fitting with the help of electronic circuitry

The scientists quite precisely described action potentials. The behavior of nerve cells and the electric circuit equation were put in relation: action potential, spike, firing. The physics of the process is exquisitely expressed; and it was many times since experimentally confirmed. The physics of the living can be experimentally tested; and these experiments can be reproduced. This kind of work eventually led to the model of the neuron meant to formally describe the natural neuron with its dendrites, soma, and axon.

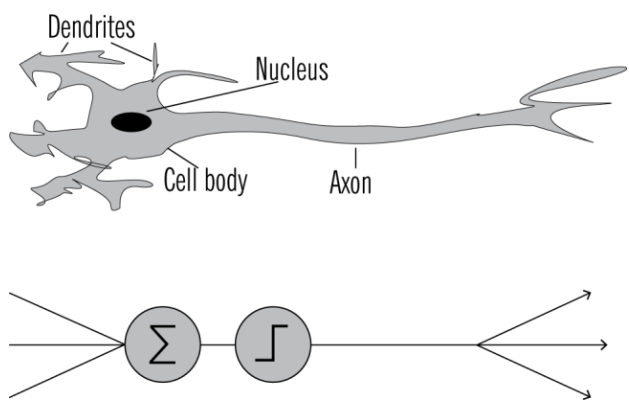

**Figure 4** From the generality of the neuron to artificial neurons

The mathematical model (of course a symbolic representation integrating iconic and indexical elements) is described through a behavior that associates inputs (such as those received by dendrites) and outputs (propagated to the dendrites of other neurons).

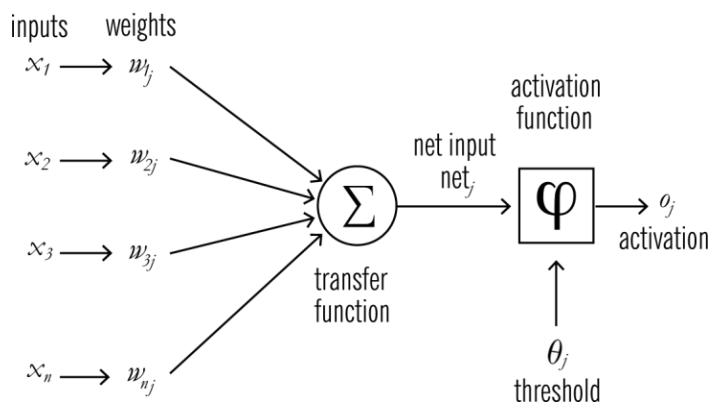

$$Y_k = \begin{matrix} m \\ (\sum w_{kj} x_j) \\ j=0 \end{matrix}$$

**Figure 5** The artificial neuron as mathematical model

The artificial neuron (AN) accepts data (supposed to be of sensory origin) and tries to provide the description of what happens when such data is processed. In the diagram (Fig. 5), there are **m** inputs to be summed up in some manner; there are $w_m$ weights, to express bias (what is more or what is less important); and a transfer function **φ**.

I focus on this premise, "Elementary, my dear Watson," for what eventually became the spectacular field of artificial neural networks (ANN), and further on of learning and deep learning, because, like the Hodgkin-Huxley model, it is representative of a machine-reductionist view that both explains its accomplishments, but also its implicit limits. The broader context involves what Shun-ichi Amari defined as the "prehistoric" period of Rashevsky and Wiener, the perceptron (debunked by Minsky and Papert), the Kohonen (1988) and Hopfield (1982) moment, the connectionist model, the first back-propagation paper (Rumelhart, McClelland, and Hinton in 1986), etc. Evidently, Hinton's work on the Boltzmann Machine (as Peter Norvig, Research Director at Google reported) and his presentations at Berkeley brought symbolic AI to a standstill.

As opposed to the physical symbol-processing paradigm, it had cognitive plausibility: the brain is always respected (even by those who don't understand what it is). Moreover there was training based on real experiences, and there was an enticing analog component: continuous representation instead of Boolean sequences. It is worth clarifying that this is not a study in history. Rather it is an attempt to take note of opportunities missed in the absence of acknowledging the fundamental distinction between the living and the non-living.

## 6 Artificial Neural Networks

If those involved in the hot topics of artificial intelligence had taken semiotics into account, they would have realized that sign processes are not abstract logical or mathematical operations. Staying within the knowledge domain associated with the neuron, we have to account for the switch from the reticular model of the brain (the reticulum being tissue formed by fused nerve cells) to that of independent and autonomous units called neurons (associated with Santiago Ramón y Cajal, but already identified to some extent before him by Johannes Purkinje and Otto Friedrich Carl Dieters). To keep this simple: from the drawings of motor neurons by Dieters to the images from Ramón y Cajal, it becomes pretty obvious that there are many types of neurons, that is, many cells not contiguous with other cells. The activity in the neuron is also remarkably rich in details in terms of the physics and chemistry at work, in particular electricity, but also in respect to the variety of its behaviors. There is nothing "mechanical," there are no repetitive patterns. Neurons seem to understand what is going on, as the eminent Gelfand observed (Arshavsky 1991). Their functioning is affected by factors different from those associated with their material embodiment. Electro-physiological recordings, which of course disturb the dynamics of the measured neuron, indicate the presence of voltage-gated ion channels on dendrites. Surprisingly, as proof of their autonomous identity, they evince action potentials back propagated from the body of the neuronal cell to the dendrites.

In a different context (Nadin 2016), I brought up the empirical observations that no two cells in the body are the same. Ergo: there are no identical neurons. Moreover, there is a continuous re-creation of each (a subject to which I shall return), at a scale defying what even Big Data is understood to be.

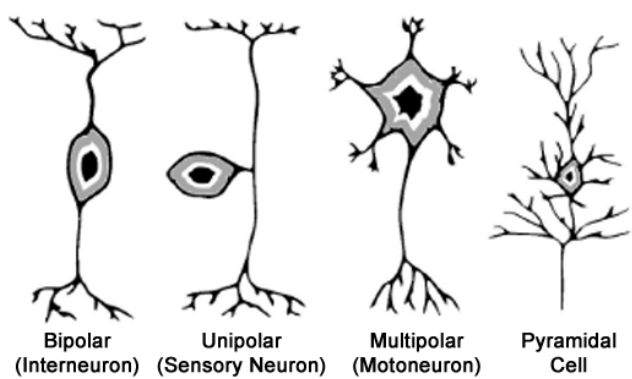

**Figure 6** Some neuron types—second level generality

The motor neuron (several feet long) extending from the base of the spine to the toe is definitely quite different from the brain neurons (Fig. 6). Whether the number is 120 billion or only 86 billion (the most recent estimate) neurons, the scale is such that the idea of replication, of numbers and variety, is at best not within our current abilities. Those who still attempt it should ask if it is the right target since it continuously changes. Synaptic activity multiplies this moving target by many orders of magnitude—well into the trillions or even quadrillions! This prompted one scientist to come up with an unusual formulation: "a number whose logarithm is itself a large number will be designated as 'immense'" (Elsasser 1996, p. 96). The logarithm of the number of synapses is very large! We will make reference to the qualifier "immense", as the argument requires (specifically, in discussing deep learning).

6.1 Learning and adaptivity

The short overview given above suggests at least two epistemological questions:

1) What kind of knowledge is afforded by an abstract model, which by its nature is less rich than the real?

2) What kinds of generalizations from the abstracted reality to reality (the artificial neuron with limited connectivity) itself are legitimate?

The AI proponents of symbol processing were not wrong in assuming that language carries information about activities within which words are formed and their use is established. That they left out everything else besides language is only a temporary limitation. In reality, they soon recognized that vision was also important—and it, too, became an AI specialization. Further on, the entire semiotics (not only words, images, sounds, but also tactility, smell, taste) can be recovered, and is in the process of being acknowledged (Peirce or not). What they missed was the understanding that within the living, causality is richer than in the physical. Learning, in a variety of forms, takes place at each level of life, the outcome being the aggregated expression of adaptivity. The artificial accommodates a subset of intelligence, if indeed self-awareness could be reached in non-living matter. The Machine Learning (ML) take on intelligence left symbols out and focused on emulating neuronal activity, understanding it only partially.

In his impressive *Reflections on a Theory of Organisms*, Elsasser (1987) gave a systematic description of the living that transcends that of Schrödinger (1951), with whom he worked. *Atom* *and Organism* (1966) goes even further, accounting for the quantum mechanics views. He tries to understand the meaning of logical complexity in order to see if it opens a better chance of understanding the living. "Individuality increases…as one rises on the evolutionary scale." That would mean that individuality, as a measure of evolutionary advance, is probably a better explanation than Darwinian adaptation. Elsasser makes it clear that "A concept of life outside or apart from a specific set of physico-chemical mechanisms is completely meaningless" (op. cit., p. 105). But he also brings powerful arguments in favor of studying the reality of the immense, non-homogenous indeterminate from a different perspective of complexity. I repeatedly called his work to the attention of the scientific community (Nadin 2003) due to its practical implications. The same applies to Robert Rosen's views. They are premises impossible to ignore by those attempting to emulate aspects of life. Rosen's work is the outcome of a meticulous architecture of thought, consisting of a mathematically founded theory of measurement and a theory of anticipatory systems (Nadin 2012). In full awareness of my debt to each of the above-mentioned scientists (and to many others), I prefer to focus here on what distinguishes artificial vs. living intelligence. More to the point: why reactive intelligence, with its associated predictive functions, and anticipation intelligence—an expression of anticipatory processes—are probably complementary, but by no means comparable, and even less reducible to reaction. The semiotic angle is that of *meaning*: reactions in the living are congruent with reactions in the non-living, and follow the cause-and-effect sequence; anticipation-informed action is the outcome of realizing the meaning of change.

**7 Awareness and Matter**

The Hodgkin-Huxley model is fully adequate for describing the reaction component of the behavior of the living. It follows the McCulloch-Pitts (actually the Rashevsky neurodynamics model of 1930, taken over by his students, probably with his blessing). It is also a blueprint for artificial products, like hearing aids or any other technological substitute for lost vision, lost tactility, etc. The abstract model of the neuron (Fig. 4 and 5)—and there are all kinds of variations to the McCulloch-Pitts representation (1943)—is no less effective in describing, broadly, what the authors termed as neuron activity, and which was actually its physics. No meaning attached, only quantities—data reaching the dendrites. Some aggregation (described as a summation of weighted values), and—*voilà*!—the output, maybe not the thought itself, but a component of it.

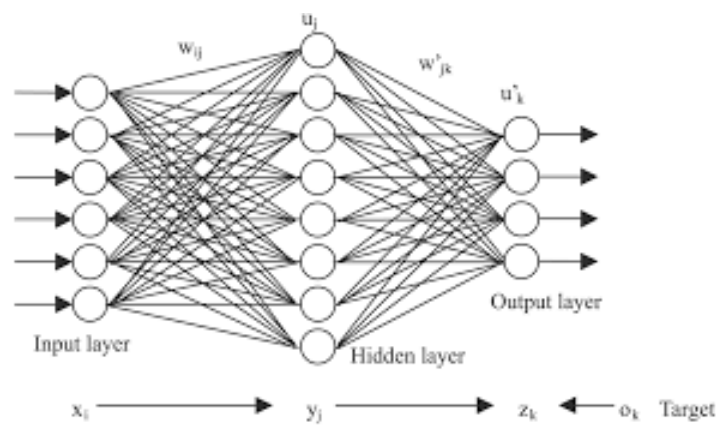

**Figure 7** Artificial Neural Networks model

Empirical evidence concerning the living, however, makes it plenty convincing that the living not only reacts, but mostly acts in an anticipatory manner. Predatory behavior, reproduction, and the ability (not just of bees, ants, and humans) to plan and establish a propitious environment for survival are easy to understand. I named only a few, since for years I have produced for everyone interested examples in abundance (Nadin 1999, 2010). This is empirical evidence of anticipation in action. Indeed, the possible future affects choices and informs activities evincing anticipatory processes. In the works of the Soviet/Russian early attempts at describing anticipatory processes[7], we find research contributions to physiology, brains science, anatomy, learning, for example, some worthy of recognition (including a Nobel Prize that decades later confirmed Beritashvili's work on the anticipatory aspects of navigation.[8]

For everything involving the vector PAST→PRESENT→FUTURE, physics and chemistry, as pertinent to any form of the living, delivers exceptionally well. For everything involving both directions, from PAST→FUTURE and PRESENT←POSSIBLE FUTURE, neither physics nor chemistry delivers. The reason for this is relatively clear: within the realm of the non-living, physics captures the form of physical law (the *nomothetic*), i.e., the repetitive nature of all processes of entangled matter and energy. In the realm of the living, creativity defines its dynamics. To avoid any confusion (and the seduction of unjustified speculation), *to create* means to make possible something that never existed before. No matter how similar something that is alive, or is an expression of life seems, it is "repetition without repetition," (an expression to which I shall return). Self-preservation of life is the fundamental characteristic of the living embodied in matter. As a self-organizing system, the living maintains its own interlocking of biological matter and energy through metabolism. Moreover, it maintains the integrity of its instantiation in a particular form of life (the individual animal, plant, insect, et.) through self-repair, for which metabolism delivers matter and energy (Rosen 1972)

These aspects of life expression suggest that there is understanding at work, i.e., intelligence. The living is not a pre-programmed machine executing commands between inception/conception and death. That which is alive makes and remakes itself continuously—a process in which genetics plays a major role (the DNA aspect). But learning (at least in what epigenetics has so far confirmed) plays probably a no less important role. Properties of living matter result from complementary processes (Fig. 8): bottom-up and top-down (Ellis 2012).

---

[7] International Conference, *Anticipation – Learning from the past. Early Soviet/Russian contributions to a science of anticipation*. Hanse Institute for Advanced Study, Delmenhorst, Germany, September 1-3, 2014.
[8] Nobel Prize for decoding brain's sense of place. Discoverers of brain's navigation system get physiology Nobel. See: http://www.nature.com/news/nobel-prize-for-decoding-brain-s-sense-of-place-1.16093

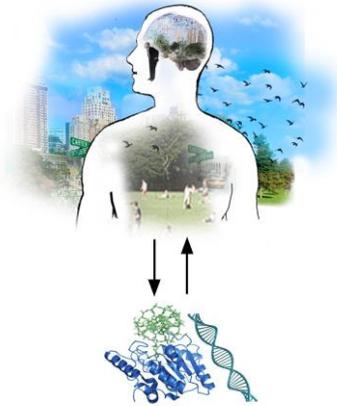

**Figure 8** From matter to mind activity and from the mind to the matter (cf. Murphy et al 2009)

## 7.1 Intelligence, Creativity, Anticipation

Biology-based considerations integrating semiotics are unavoidable. They are absent from the work of those who conceptually address the possibility of artificial intelligence, and, moreover, absent from the considerations of the proponents of computation and its expression as AI are. The intelligence of the cell, or the intelligence of connected neurons, not to mention the intelligence of each organ, is not reducible to symbol processing. But neither those whom Dreyfus and Weizenbaum addressed, nor those who felt attacked were eager to invest in understanding what it takes to survive. The never-ending change of any and all living entities entails creative processes. Reproduction (sexual or asexual) is, from among a large variety of creative processes, the most prominent. Self-preservation guides variation and selection, from the cellular level to that of the species. It succeeds to the extent to which anticipatory processes lead to successful action.

The *vitalist* distinction between what is alive and what is not was compromised for good. It prevented systematic attempts to understand what defines life because such attempts were qualified as vitalism. Of course, when scientists of unquestionable performance (theoretical and experimental) frame their object of interest as intelligence in something other than the living, the label "artificial" is a statement of epistemological consequence. They indirectly ascertain that there is something—the artificial—that is not of the same nature as the living. For example, Hotchkiss (1958, p. 129) believed that "Life is the repetitive production of ordered heterogeneity," evidently missing the lack of repetitivity: "repetition without repetition" (Bernstein 1967).[9] In the spirit of science, the scholars of the artificial wonder what aspect of the living could be emulated outside the living in order to achieve what they described as intelligence. The creative, non-repetitive remaking of life? Heterogeneity? No stone is suspected of intelligence during its long existence in a shape in which it is possible to guess the future pebbles or grains of sand. It is also possible to calculate the energy needed for the "mill" of time—heat, humidity, pressure, wind, etc.—to grind the stone. However, the behavior of even the most insignificant living entity qualifying as intelligent cannot be predicted. This places the description of intelligence discussed at the Dartmouth Conference (and since then in a never-ending series of conferences) or discussed by Dreyfus and Weizenbaum in a different context. Learning, evinced in the way behavior is affected by experience, is probably the common denominator—if indeed the artificial can learn. In the absence of an effective distinction between reaction and anticipation, nothing else is really definitory in respect to intelligence. Nobody has suggested that AI should emulate survival, or ensure preservation of life (although in our days, some very visible proponents of intelligent technologies, such as Kurzweil, are becoming addicted to the hope of immortality, or at least an extended lifespan). Nevertheless, it is impossible to ignore a simple empirical observation: Preservation of life is what defines the intelligence of the living (and not the so-called IQ that has almost nothing to do with intelligence). Change in living matter is existential. It leaves traces that eventually form knowledge (no matter how limited) of self and of the world in which the living unfolds.

---

[9] Н.А. Бернштейн. Физиология движений и активность, 1990

At Dartmouth, the subject was not a synthesized intelligence mimicking the intelligence of the lifeless moon or some planet, or the intelligence of a stone. But if it were only the intelligence it takes to win a game (chess or any other), it missed the most important aspect: the creation of the game itself, as one of many instances in which human beings shape their own condition. The game of chess documents learning, the ability to represent and to make associations, the understanding of reward, the awareness of aesthetic expression—and much more. This is yet another example in which the laws of physics and knowledge of chemistry leave the "Why?" (of the invention of the game of chess, and of playing it) question unanswered. The AI proponents had a broader view, transcending checkers and chess: Can we find an effective procedure for, let's say, diagnosing (disease or device malfunction), distinguishing between desired and less than desired outcomes of actions, between good and bad plans, etc.?

From an epistemological (and even logical) perspective, there is no need for intelligence in the lifeless. For an observer, the nature of interactions in the lifeless is describable in quantitative terms. As we know, such descriptions underlie the experiment through which new explanations can be tested. The nature of interactions in the living are only partially described in quantitative expression. The necessary condition of life preservation constitutes the domain of the meaning.

Semiotics, including symbolic representation (as well as iconic and indexical), but mostly ascertaining endless interpretation (i.e., semiosis) is complementary to the quantitative. This was entirely missed in the disputes of those days, as it is absent even more in our days. That which is alive stays alive because it is capable of understanding; that is, it interprets. It is not intelligence that made computer programs for playing chess beat Dreyfus or Kasparov—IBM itself confirmed this. Even the idea of subjecting a living player to the chess program is absurd: "If the bullets don't kill you, you deserve to live" would be equivalent to gladiatorial combat through which a slave would either die or be freed, but not a proof of intelligence (except in Matrix, since religion always expands into culture).

## 8 G-complexity Revisited

IBM was pretty darn honest: "Deep Blue relies more on computational power and a simpler search and evaluation function."[10] Shannon (1950), yet another of the great minds present (though marginally) at Dartmouth, calculated the lower bound of the complicated game-tree of chess (estimated at $10^{43}$); others (such as Allis 1994) calculated the upper bound. But after defeating (in 1989) an earlier IBM chess machine, Kasparov was more precise: "Chess gives us a chance to compare brute force with our abilities." One hundred years of grandmaster games form a large body of knowledge that the chess program can rapidly access. Specialized hardware and sophisticated data processing are part of the very broad picture—again, both in the allowed operations and in the ultimate goal. It is neither so simple as to be trivial nor too difficult for satisfactory solution. There is no learning, there is no intelligence, even though the proponents of AI would have a tough time admitting it. IBM wants to sell technology, not theoretical assertions. What is and remains the focus is the machine and the associated "theology" it has propagated and perpetuates. Before addressing this aspect, one more remark: the word impossible to avoid in reading the volumes dedicated to AI (in support or critical of it) is *complexity*.

---

[10] Does Deep Blue use artificial intelligence? https://www.research.ibm.com/deepblue/meet/html/d.3.3a.ht

In previous publications (Nadin 2013, 2015, 2017a) I argued that this very important concept loses its epistemological significance when used arbitrarily. There is no complexity in the game of chess, as there is no intelligence in having an effective procedure (remember Hilbert?), or what has by now become an algorithm, defeat a human player, disadvantaged by a machine' sheer number-crunching power. Gödel produced the proof that a certain formal system (all going back to Hilbert) is not decidable; that is, it cannot be fully and consistently described. Chess is decidable; so is Go; and so are many of the newest targets of AI masquerading as machine learning (deep, deeper, etc.). Indeed, as I generalized from Gödel to what I call *G-complexity* (Nadin 2014) it became clear that the living is G-complex, i.e., not decidable, while the domain of physics—from the simpler aspects of movement (as change in position) to the more complicated, including the physics of measuring gravitational waves associated with the coming together of the universe, is decidable[11]. Within a unified systems perspective, observables over states of the system form the starting point. Therefore, let us consider the phase space of physical or chemical processes and compare them to the phase space of living processes—a great suggestion from Giuseppe Longo (2013).

In the non-living, mapping from states to numbers captures the nature of change as quantitative, i.e., subject to measurement. In the living, mapping to numbers only partially describes the nature of change, especially as a consequence of the empirical fact (knowledge acquired through observation over a well-defined duration interval) that the observables making up the phase space continuously change. Therefore, to account for the dynamics of life, it becomes necessary to perform mappings from states to meaning (as the parameter of change) relates to the self-preservation of life. In practical terms, this suggests the need to generate sequences of maps (observables over the various parts of the continuum of life, such as early existence, childhood, maturity, etc.), and to examine, for each of the variables, their change, their relation to previous and possible future states. The number and variety of parameters describing the non-living is finite (no matter how numerous they can be). Interactions in lifeless matter and among non-living entities are described by the dynamics of action-reaction, i.e., deterministic causality (including, for instance, processes described in chaos theory, i.e., dynamic systems). Inferences from parts—a sample from a stone, a liquid, or a gas, etc.—are possible and effective because interactions through which matter and energy are interlocked are preserved (up to a certain scale). Variations (an expression of imperfect descriptions or measurements) average out. Of course, the finer the granularity of observations or measurements, the higher the possibility that what is measured might be noise, not the phenomenon as such. One thing is sure: the dynamics of lifeless matter is fully and most of the time consistently described through the variables relating the past to the present. And one more thing is evident: there is no intelligence at work in physico-chemical processes.

---

[11] The 2017 Nobel Prize in Physics was awarded for decisive contributions to the LIGO detector and the observation of gravitational waves.

Intelligence emerges and resides in the living. It is a necessary characteristic of it. It cannot be detached from it. Given the action of self-preservation of life, the dynamics of the living cannot be described and explained without considering the possible future—which obviously includes death. The number of variables describing the dynamics of the living is as open-ended as the possible-future-based choices it faces as it unfolds, in an individualized manner, over its viability interval. The interlocking of energy and matter in the living makes possible the simultaneous condition of *sameness* (in species, in offspring) and *difference*, expressed as irreducible individuality, of which lifeless matter has none. Inferences from parts to whole, fully possible in the decidable (the physico-chemical) are misleading in the living. Interactions through which living matter and energy are interlocked is specific to every life level: cells, membranes, tissues, organism, etc. There is intelligence at each level because self-preservation of life implies awareness and the whole of life.

8.1 Playing god

Indeed, intelligence is not the outcome of a repeatable effective procedure, but rather an expression of adaptive behavior. Lifeless matter is homogenous; atoms, molecules, and aggregates are of the same nature. Life embodied in matter is heterogenous, from the cell level to tissues to organs, up to the organisms. This heterogeneity is reflected in in the undecidable expression of intelligence: never fully describable, never consistent (the same action can be interpreted by an observer as intelligent or lacking intelligence).

It was childish to speak (repeatedly) and write about what machines cannot do, and later about what machines *still* cannot do. The media liked it. Even those who do not reject science and technology found the examples convincing. Pamela McCorduck remembers in her e-mail:

*My last interaction with him [Dreyfus] was on a San Francisco radio show. I was doing publicity for a reprint of Machines Who Think [2004] and he called in to*

*mock AI because the self-driving competition in the previous year had run off the road after five miles.*

The philosophic arguments obviously mean nothing when the focus shifts to false arguments, such as: "There is nothing that a human mind can conceive that cannot be imitated." Absence of this understanding leads to the violation of the most important commandment of science: "Don't play God!" by attributing your own convictions, usually justified on high principles, to the higher authority, which you yourself conceived.

Remember Isaiah realizing how foolish it is to bow to gods we ourselves created. The various "cults" and "churches" of the theology of determinism morphed into the machine religion. From the "stable" of authors represented by agent-priests and publishers of newer and newer versions of the same come headlines such as "We are in the Presence of a Formidable Creature"[12] associating artificial intelligence with the oriental mythical creature Djinn. In the absence of a better explanation of why intelligence is not the outcome of calculations only, the call for measuring more ("Shut up and measure") without knowing why, sounds like a religious precept  more than a scientific principle. The theology of reductionism pushes the authority of the god it made up, not realizing that authority itself is a boundary condition.

Playing god has a history about as long as that of the human being. One of its most conspicuous forms is machine reductionism. We have seen that intelligence, i.e., understanding what it takes to perform actions through which self-preservation of life takes place, is associated with learning—but is not reducible to it.

---

[12] Reproduced on Edge.org a report from the Süddeutsche Zeitung, Oct. 13, 2017.

## 8.2 Extending human capabilities

The drive and need to extend human capabilities informs tool-making (Fig. 8). For the longest segment of humankind's history, the focus was on physical abilities. Hunting and gathering, as well as, later on, agriculture, benefit from augmenting the application of the force of muscles. Eventually manufacture (making things by hand), the trades, and industrial activities led to more tools, and even to the expectation of automating their use. That's how machines emerge and change the nature of work.

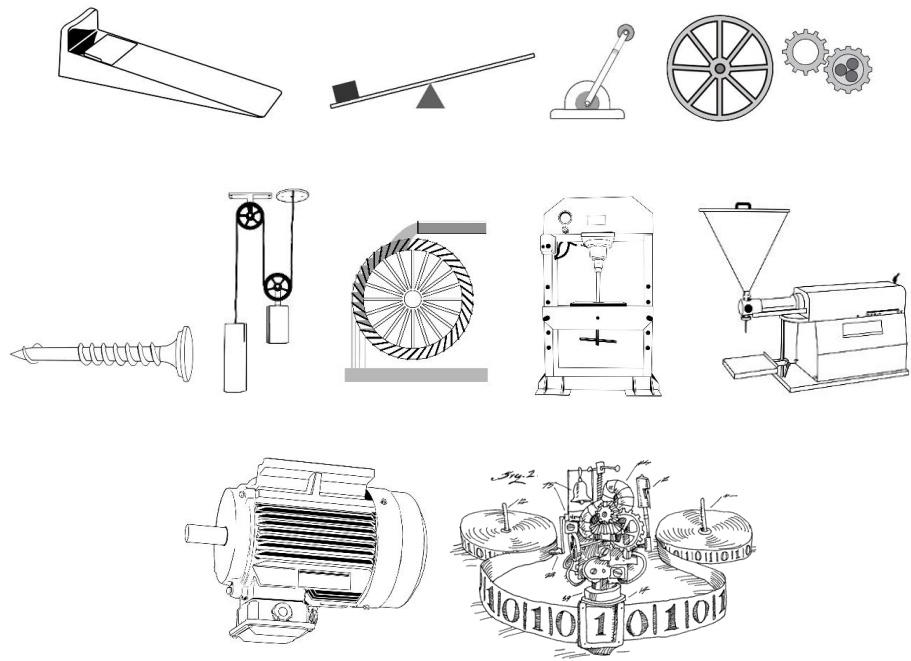

**Figure 9** From tools extending physical abilities to machines—inventions and not discoveries

In the process of conceiving tools—which is an expression of intelligence, i.e., understanding and learning embodied in the tool—intelligence itself changes. Those who tried to emulate it in the artificial missed the opportunity to acknowledge such changes. Adaptive processes notwithstanding, the creative act of tool-making is consubstantial with the re-creation of the human being itself. Far from being passive embodiments, tools embody knowledge of oneself and of the activity. The making and use of tools trigger an open-ended cycle of adaptive processes. Individuals using a hammer, for instance, are physically and also mentally "enhanced," That is, their intelligence changes. There is nothing exceptional in the observation that the human being might see in the tool more than what he or she contributed to it, more than an expression of human abilities. Even in our days, one asks, "How come…?" (…a lever helps to move a heavy object or helps to lift it). The answer today is in the straightforward physics of the tool, and in the simple mathematics describing that physics. Together they make engineering possible. This answer, associated with actions and objects chosen, was derived inductively, one application after the other, not deductively, and even less axiomatically. The inductive nature of activities leading to tools, invented through trial and error, does not preclude the forming of explanations. But such explanations have nothing to do with the particular practical experience. As opposed to the engine as an extension of the individual, Big Data-based computation and the Turing machine are disconnected from the experience. So is the attempt to emulate intelligence and the processing of symbols, or deep learning connectionism and the perception of connectedness.

The conception and making of tools is part of the human being's self-constitution through what we do. In this self-constitution, there are rational elements (choices made, tested, validated) and irrational elements: "The tool does it because…." (Here we can fill in anything we can think of, such as "I am a better person," "I am lucky," "There is some force out there that does it," "The spirit moves it"). The limits of our knowledge are not in words or language. (Remember how the rush to capture the Higgs boson at CERN forced a huge explanatory effort because the Standard Model of particle physics did not have an appropriate language for explaining it.) The limits of knowledge are in our activity—to which words and/or language often belong. The hypostatized tool—the CERN accelerator—is yet another instance of missing the connection between what we do and the outcome of our activity There was no "God" particle to find, but only what scientists predicted, based on a given model of the physical world and the forces at work in this world. For many, including scientists of high repute, it appeared as though the tool was magical—just as humans have considered tools of all kind throughout their history. It seems that the quest for higher authority is not as much a symptom of the beginnings (of humankind facing reality while unable to understand it), but one of implicit limitations otherwise difficult to deal with.

Machines emerge as the self-constitution of the human beings enables and requires—for reasons of life self-preservation—higher efficiency. Tools "extended" arms and legs, and "made" muscles seem more powerful on account of energy spent by the person using them. When energy other than that of the human being, or of oxen or some other animals, is used in order to augment efficiency, the result is the machine (the engine) that does what one or several humans would have done, but with energy from outside (hydraulic seems to come first). If the tool as an extension of physical attributes was hypostatized, the machine, using the invisible energy of what moves it, "gives it life," becomes one of the gods, and in the long term, the god of determinism.

The clock (behind which, in its pendulum embodiment, one identifies gravity) impressed upon Descartes (and others) reductionist views. The machine that humans made, the engineered, has become the model for those who made it (like the gods of religion).[13] One can imagine how more subtle machines, such as computers and artificial neurons, suggest associations that extend so far as to form religions and sects of the machine. The mechanical procedure that Hilbert asked about in regard to proving mathematical statements is by now a logical procedure. Not unlike the hypostatized machines of the past, it becomes a) an explanation for how the human being operates; b) in particular how the mind or brain (depending on which faction one upholds) operates; c) the model of it.

---

[13] "Who but a fool would make his own god— an idol that cannot help him one bit? …the wood-carver... uses part of the wood to make a fire. With it he warms himself and bakes his bread. Then—yes, it's true—he takes the rest of it and makes himself a god to worship! He makes an idol and bows down in front of it! … he takes what's left and makes his god: a carved idol! He falls down in front of it, worshiping and praying to it. "Rescue me!" he says. "You are my god!" […] Such stupidity and ignorance! […] The poor, deluded fool … trusts something that can't help him at all. Yet he cannot bring himself to ask, "Is this idol that I'm holding in my hand a lie?" Excerpts from Isaiah 44:10-18

"The brain is a computer," has been stated in many ways by too many (with high Google scores) to be simply ignored. (This actually means that a subset of reality, i.e., the brain, is the outcome of computation.) This machine is supposed to process symbols (as was already pointed out here, they are not really well-defined) like the living do. In reality the digital machine is a construct, an embodiment of a logic (Boolean) applied to a language with a vocabulary consisting of 2 letters. (Other types of machines might have a different structure.) This fact does not preclude the view that it is equivalent to a human being, but has even led (as we have seen) to the notion that reality is the outcome of a larger computation. McCarthy (in a text co-signed with Pat Hayes, 1969, p. 5) went even further in the direction of machine idolatry. "The physical world exists and already contains some intelligent machines called people." Let's repeat: "intelligent machines called people." If so, why AI?

8.3 Why AI?

Indeed, Wiezenbaum stated that, "Since we can all learn to imitate universal Turing machines, we are by definition universal Turing machines ourselves." He added, "That is, we are *at least* [sic!] universal Turing machines." The constitution of a universal church based on faith in computation, on algorithms in particular, is the result of ignoring the second part of Weizenbaum's assertion: "…we are at least universal Turing machines" (1975, p. 71). By this he meant not the "physically embodied machines, whose ultimate goal is to transcend energy or deliver power," but rather the "abstract machines that exist only as ideas" (which is the case of the Turing machine).

Of course, no one would question the accomplishments associated with computation. Practically everyone in the world either uses computers or is affected by their ubiquitous presence. By now everything is either seen as an outcome of computation or will soon be computerized—or abandoned. Intelligence is only the most provocative aspect. That a decidable entity—the Turing machine—is equated with the undecidable brain, or for that matter with any other part of the living, is not only accepted on faith—indeed, there is no evidence for this—but promulgated as the only acceptable expression of science. Once again, the premise, i.e., the famous Hilbert challenge, is relegated to the drawer of insignificance. Gödel's principles, as well as Turing's proof are used to justify the opposite of what each of them asserted. Let us recall that Lucas (1961) attempted to demonstrate that Gödel' theorem refuted mechanism, but Lucas's arguments were ridiculed (and in the meantime forgotten).

There are some additional arguments to be considered when examining the "church of the machine" and the "cult of the algorithm." In its Turing machine embodiment, quite different from the other possible machines that Turing considered, a purely syntactical procedure affords the execution of programs. There is nothing else to it but data, and as such, algorithmic computation, spectacular in many of its applications, is the most effective data processing procedure invented so far.

**9 The a-Turing Machine Is Only One Among Others Possible**

With the Turing machine, the real beginning of automated calculation was reached. Behind his theoretic machine lies the problem of the possibility of an automatic testing of mathematical statements. Hilbert was convinced that mechanical calculations were the basis for them. The meta-level of the enterprise is very relevant:

a) objects in the reality of existence → representations → acts upon representations → new knowledge inferred from representations

b) objects → numbers → counting → measurement → ideas about objects → ideas about ideas

Hilbert's conjecture that mathematical theories from propositional calculus could be decided—*Entscheidung* is the German for *decision*, as in proven true-or-false—by logical methods performed automatically was rejected.

The consensus is clear: Turing provided the mathematical proof that machines cannot do what mathematicians perform as a matter of routine: developing mathematical statements and validating them. No less important is the insight into what machines can do, which we gain from Turing's analysis. Recalling a conversation with Turing (in 1947), Wittgenstein wrote (1980): " 'Turing's machines': these machines are humans who calculate. And one might express what he says also in the form of games." Turing (1948) also gave a description: "A man provided with paper and pencil and rubber, and subject to strict discipline, is in effect a universal machine." At a different juncture, he added: "disciplined but unintelligent" (1951). Gödel would add, "mind, in its use, is not static, but constantly developing" (1972). "Strict discipline" means: *following instructions*. Instructions are what the algorithm, the effective procedure, is. In contrast, intelligence at work often means shortcuts, new ways for performing an operation, even a possible wrong decision. Therefore, non-algorithmic means not subject to pre-defined rules, but rather discovered as the process advances, are part of intelligent performance.

Automatic machines (*a-machines* as Turing labeled them) can carry out any computation that is based on complete instructions. The machine's behavior is pre-determined. It also depends on the time context: whatever can be fully described as a function of something else with a limited amount of representations (numbers, words, etc.) can be 'measured', i.e., completed on an algorithmic machine. The algorithm is the description (the "recipe").

With the *a-machine*, a new science is established: the knowledge domain of decidable descriptions of problems. In some sense, the *a-machine* is no more than the embodiment of a physics-based view of all there is.

Turing knew better than his followers. In the same 1951 paper, Turing suggested different kinds of computation (without providing details). Choice machines, i.e., *c-machines*, involve the action of an external operator. Even less defined is the *o-machine* (the oracle machine advanced in 1939), which is endowed with the ability to query an external entity while executing its operations. The *c-machine* entrusts the human being with the ability to interact on-the-fly with a computation process (as, for example, in supervised learning neural networks). The *o-machine* is rather something like a knowledge base, a set subject to queries, and thus used to validate the computation in progress. Turing insisted that the oracle is not a machine; therefore the oracle's dynamics is associated with sets. Through the *c-machine* and the *o-machine*, the reductionist *a-machine* is opened up. Interactions are made possible—some interactions with a living agent, others with a knowledge representation limited to its semantic dimension. Predictive computation is attained; *anticipation becomes possible*.

9.1 The other Turing machines

The theoretic construct known as the Turing machine—in it's a-, c-, and o- embodiments—will eventually become a machine proper within the ambitious Automatic Computing Engine (ACE) project. (In the USA, the ENIAC at the University of Pennsylvania and the IAS at Princeton University are its equivalents.) "When any particular problem has to be handled, appropriate instructions…are stored in the memory…and the machine is 'set up' for carrying out the computation," (Turing 1986). Furthermore, Turing diversifies the family of his machines with the *n-machine*, unorganized machine (of two different types), leading to what is known today as neural networks computation (the B-type *n-machines* having a finite number of neurons), which is different in nature from the algorithmic machine.

Von Neumann (who contributed not only to the architecture of the Turing machine-based computer, but also to the neural networks processing of data) asserted that, "…everything that can be described with a finite number of words, could be represented using a neural network" (Siegelmann and Sontag 1991). This is part of the longer subject of the Turing completeness or recurrent neural nets. Coupled Turing machines, networks of Turing machines, oracles via quantum randomness, and infinite time Turing machines are extensions impossible to ignore because they are the outcome of new questions regarding the nature of computation. Let it be noted again that interactive computations is not reducible to Turing algorithmic processing. External input, through which interactivity is obtained, cannot be modeled by a Turing machine. (Without going into details: the tape on the Turing machine is supposed to have all input available from the start. Interactivity undermines the fulfillment of this condition. Interactive computation opens up the possibility of semiotic grounding. Meaning becomes part of the computation through the process of interaction.

9.2 Intelligence vs. data processing

With all this information in mind, ignored as much as semiotics was in the years preceding and following the Dartmouth Conference, we return to yet another ambiguity that marred the conversation about computers. Computers in the a-machine embodiment process data. Intelligence is different from data processing, even within the scope of what was described as physical symbol processing. The premise for intelligence is the *understanding* of what it takes to address a question leading to some action or to no action. Understanding conjures information. Data becomes information once it is associated with meaning. Wheeler (1989) exemplified this in interpreting radioactive decay: the click on the Geiger counter makes sense if we reference it to the process it documents: the atom has decayed (Davies 2004). This is an a-causal process. Under "participatory universe," Wheeler understood the epistemological universe in which we do not just reflect what we encounter in the world, but we especially contribute meaning to the perception. Subjects co-constitute what is, including their own being in the world. The sound of the heartbeat is another example. The sound as such can be captured as data (frequency, intensity, spectrum). More important is the question: What does it say? (What does it mean?) A machine listening to the heartbeat without referencing it to the cardiovascular reduces the process to the physics of sound generation and propagation. A good cardiologist seeks the meaning.

In the absence of a semantic dimension, computation enlisted the heroic effort of ontology engineers, who "translate" encyclopedias (like Britannica or Wikipedia) into the language of computers. This pseudo-semantic dimension (based on descriptions using first-order logic) still cannot supplant what semiotics would afford: the pragmatic dimension (Fig. 10).

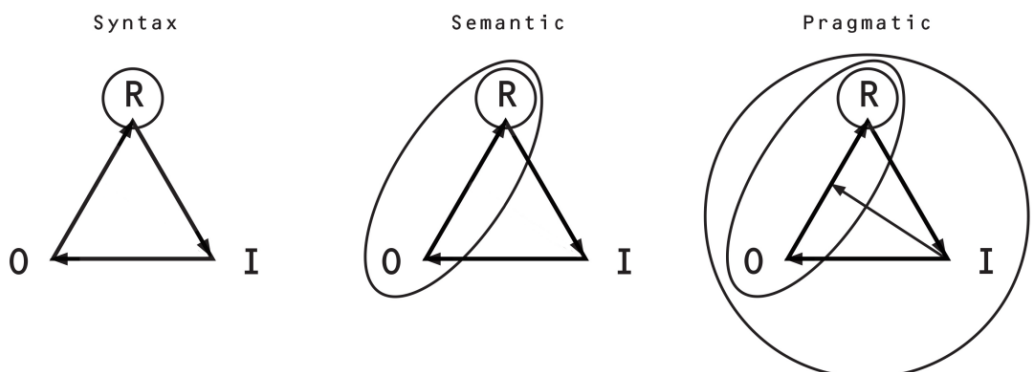

Defining syntax, semantic, pragmatic (in connection with Peirce's sign definition)

**Figure 10** Syntax (formal aspects of representations), semantics (link between representation and the represented), pragmatics (sign interpretation process through various activities involving representations).

Intelligence, itself a human construct, an abstraction, is supposed to describe what is needed to attain a goal. Human performance is not the automatic consequence of the nomothetic, i.e., of the laws describing the dynamics of the world. Even the deployment of tools, never mind their invention, proves this. There is no intelligence at work in the fact that the living and, for that matter, the non-living fall down, not up. Gravity explains the direction of falling, and physics accounts for its details (the laws, i.e., the nomothetic). However, there is intelligence at work, in implicit or explicit expression, in "falling the right way" that is, in such a manner as to prevent harm or minimize the consequences of falling. Intelligence does not change the laws of physics, but is conducive to discovering such laws and to informing a behavior corresponding to explicit or experiential awareness of physics. Learning is the process through which this takes place. Learning covers awareness of the nomothetic aspects of life, but also the ideographic aspects, i.e., the perception of uniqueness. There is creativity at work in human expression, regardless of whether it is the formulation of an idea or saving seeds, nailing two planks together, or finding a new path towards a mountain peak.

There is more, much more to the dynamics of change than falling. The entire gamut of motion—how the living moves in the process of self-preservation of life—is part of the same awareness. So is awareness of chemistry—for instance, what can or should be eaten, inhaled, drunk, used for cleaning—or what should be avoided, or what is missing. This shorthand description of intelligence at work in the living documents the empirical observation that it is pragmatically driven. No experiment can prove it or test it. The historic record, similar to that of evolution, carries the meaning with it. The syntax of all the sequences through which actions are performed partially explains the outcomes: a wrong move and one gets hurt; a wrong substance in the food chain can undermine self-preservation of life. The semantic dimension adds to the understanding of choices: What does it mean to fall on ice or in the water as it applies to motoric expression, or to other choices in the physico-chemical realm? The pragmatics integrates all levels.

## 9 Timing vs Clocking

Definitory is the Why? question, the goal for whose attainment  all living resources are deployed. Intelligence is evident after, not before. It is not a tool box, but rather a process of enlisting available resources. We'd better take this understanding with us as we return to the attempts made so far to define artificial intelligence. Examining the outcome of human expression, we realize that the intelligence of the living is purposeful, or better yet, goal-driven. Pseudo-semantics made available through contorted ontology engineering facilitates the "understanding" of speech by computers, i.e., the fast referencing from a word (e.g., egg or dog) used in a conversation to the encyclopedia. The "understanding" of images, or of sounds (via machine learning) is also facilitated through digital definition (e.g., "This is a stone, not a…"). In the absence of a pragmatics-driven process, through which representations—not only symbols—support anticipatory processes that result in actions, successful or not, there is no intelligence to account for. In simpler terms: the premise of intelligence is the understanding of means (the "is" situation) and purpose (the possible future). Often in the living, intelligence is implicit in the process: the grip on the hammer (which even Merleau-Ponty was aware of) or on the handle of the cup of water is anticipatory. But the outcome is by no means monotonic: the same activity might be successful—lifting a full cup without spilling—or not.

The program at the Dartmouth Conference—and its continuation in a variety of meetings, publications, experiments, etc.—missed the necessary condition of intelligence, which is the understanding of what is to be achieved and of the means used. If it had been formulated as "Automation of tasks associated with intelligence," no one could have objected to it. It was not intended as deception. I do not question the integrity of those involved, but rather their premises. Everyone present believed that intelligence can be obtained in the non-living artificial. Upon further examination, the Dartmouth blueprint falls within the theology of determinism, promoting the language and views of Descartes, who built upon Plato's "Nothing can come without a cause" (*Timaeus*). The cause would be intelligence, which as we have pointed out is by no means a given, but rather an expression of the adequacy of aggregating available resources. In the deterministic perspective, acknowledging a possible future as part of a broader understanding of causality, that is, beyond the cause-and-effect model, is impossible. Nobody is born intelligent (psychology's fallible and meaningless IQ model ignores this). Everyone achieves intelligence in the process of self-constitution through what one does.

I am not prepared to dispute machines' outcome, and even less to set limits (what they can or cannot achieve). Nevertheless, it escapes my understanding why the confrontation was focused on performance—automate the playing of checkers (Schaeffer et al 2007), for instance—and not really on intelligence as an expression of the creative nature of all life processes. Of course, the renewal of cells in the body, at various rhythms, within the timespan of life is probably the most powerful example of living intelligence resulting in creativity. The outcome is the self-preservation of life, pragmatically expressed through performance in any and all activities, including the invention and playing of chess, checkers, Go, or any other game. But I prefer to take another example here, because it brings into the discussion the notion of rhythm, and thus that of duration and time, impossible to ignore when posing questions regarding intelligence.

Bernstein (1967) provided empirical evidence regarding the co-variation of the elements engaged in human motion. The neuromotor system, quite surprisingly, consists of more elements than what a machine would require to execute the same operation. Moreover, these components seem to have a different rhythm in their activity, almost as though different clocks are at work. Yet, consistency across tasks is accomplished. The interaction among joints, digits, muscles, sensory units, etc. suggests a very interesting configuration: centralized (with the brain as center) and decentralized coordination, hierarchical and non-hierarchical modes in which local intelligence—i.e., understanding and central intelligence are interrelated. A moment of force in a joint crossed by several muscles is reached through many possible configurations. There is a state variability and there are many trajectories leading to what was described as "bliss of motor abundance" (Latash 2012). Concretely, changes are unique expressions, best described by what

Bernstein called "repetition without repetition." It is clearly different from the repetition cycles defining the physical (from the astrophysical to the microlevel). You need one duration machine, i.e., the clock, to account for strict repetition and to predict events within the repetitive patterns of physics: the next eclipse, the behavior of rockets in outer space, the breakdown of a bridge or an engine. You need means different from the clock to describe variable rhythms in the living. Duration as "a number of change" (*arithmos kineseos*) in respect to the before and after (*the proteion* and the *hustero*n, as Aristotle formulated it) is pertinent to movement, which is, of course, a form of change. Nevertheless, duration is established after the change, not normatively (as in the functioning of a machine) before. Duration of similar movements is variable on account of the integrated nature of the living (holistic entity). Quantitative descriptions of non-living matter dynamics (rate of change) are appropriate because the non-living physical clock is of the same nature as the change it helps characterize. This is not the case with the living. All that such a clock could help describe is the sequence of durations pertinent to the physical aspect of any form of existence. The clock, pertinent to physical phenomena, returns, at best, a record of duration. Probably rhythm would be a more appropriate way to describe timing in the unfolding of life.

The "clocks that are not clocks" in the living have variable rhythms: matter is influenced by the mind. Time flies when we're having fun; time freezes when we are tormented.

## 10 Intelligence is Possible only Over the Threshold of G-Complexity

To seek intelligence that can readily play chess, navigate, diagnose (disease or even some mechanical or electrical system malfunction), learn math or a language is to set a static target, and declare success on account of having reached the desired goal: winning a game, reaching a destination (such as in GPS-based navigation), issuing a diagnosis, etc. The mechanistic view of intelligence assumes well-defined targets. The navigation system guides the driver (or an autonomous vehicle) towards the identified destination. About open-ended choices: How do I get a nice view of the Eiffel Tower? Or: What about the bridge that just got washed away by a storm or some other event? Checking the best database of skin conditions can be helpful in diagnosing a skin condition. But what about new skin conditions, associated with behaviors impossible to index because they pertain more to possible futures than to the statistics of what already happened? The thermostat, discussed at the Dartmouth Conference, unintelligently automates temperature control. However, thermoregulation in the living requires a different understanding. Thermal receptors in the skin and throughout the body are distributed; neuronal activity also extends throughout the entire person. There is, of course, central coordination (via the central nervous system), but not to the extent of getting the same temperature in the whole body—some body parts are kept cooler than others. Shivering triggers warmth. Sweating increases heat dissipation. Intelligence is at work at each level of the organism, whether in the thermo-regulation or in any other form of self-control. The target changes; intelligence provides means for adaptive behavior.

We have here examples of the concrete instantiation of what G-complexity ascertains: There is a threshold between the living and the non-living above which intelligence is manifest, and below which it is impossible (or can be mimicked, at best). To rely on the laws of physics, or on the model of forces that would explain everything within a unified field of physics, is epistemological suicide. No wonder Swift made fun of those considering the machine that might write books on philosophy, poetry, politics, laws, mathematics, and theology (Peirce 1887). Without pursuing the distinction here, we make reference to the almost universally accepted mechanistic view of homeostasis—projecting a machine understanding of how the organism works (Cannon 1932, 1945)—and the anticipatory view of allostasis (Sterling and Eyer 1988), involving feed-forward processes. Allostasis captures the dynamics of possible future events ahead of their real influence as they become actual. Even when you declare that some part of the organism (e.g., the brain) is a machine (a computer), or that the organism behaves like a machine (homeostasis), the living does not miraculously change its undecidable condition only to align with the religion that inspired the reductionist view. The tortured logic of deterministic theology cannot "turn water into wine." The living remains undecidable; anticipatory action, couched in complexity, is definitory.

## 10.1 "Inherent impotencies"

On this note, we can revisit the machine view as such. What Peirce defined as the "two inherent impotencies" of every machine not only deserves to be recalled, but also given our attention:

> In the first place, it is destitute of all originality, of all initiative. It cannot find its own problems; it cannot feed itself. It cannot direct itself between different possible procedures. [...] In the second place: it has been contrived to do a certain thing, and it can do nothing else. (Peirce 1887)

First an observation of principle: The matter-energy interlocking, as pertinent to a fundamental law of physics and chemistry, makes us aware of the fact that everything experienced in culture is at the same time what it is (let's say a book, an idea, a machine, a game) and the historic record, i.e., what it took to make it, to become what we experience. It took matter and energy, of course, but it also took interactions through which the living manifests itself as self-preserving its life. If instead of practicing the religion of determinism we were to apply the rationality on whose basis the cause-and-effect distinction was made, we could understand the machine from this perspective as well.

What does this mean? We will not be able to replicate any characteristic of the living, not even its reactive component, without spending the energy and engaging the interlocked matter that made it possible in the first place. Surrogates do not come cost free. To make something out of nothing, and to make it quasi-instantaneous—although it took a long time to become what it is—qualifies as magic—which almost all religions lay claim to. We shall see why not only a knee or hip replacement (within the spare-parts medical notion of the human as a machine) is energy expensive and intensive, but even in "playing" checkers or chess, the automation comes at a high cost.

Machine learning discovered this through trial and error. Speech recognition, into which ontology engineering and neural networks converge, has behind it training on many years of data. Plus: this data was also transcribed—at a huge, but unavoidable cost. What is missing before we can employ the label "intelligent" is the realization that for the living, "We know more than what we learned" (Nadin 2003), and even Polany's thought, "We know more than we can tell," are indicative of life itself as learning, not a function added. That is, there is implicit knowledge at play in the living, coming from interactions, but not available in the artificial (where interaction is physical cf. Newton's laws). To assume that what we know came exclusively from outside the knowing subject—as is the case with any machine—leads to explanations that cannot exclude the magical. A great deal of living knowledge is generated from inside, because the living, as the state of being alive (process, not outcome), integrates the individual in the world. Cognitive activity aggregates data from all others parts of the living subject and facilitates associations.

In articulating these thoughts, I am aware of how AI, as physical symbol processing, and its negation through machine learning, evolved. Of course, AI practitioners of the symbolic processing initial steps are by now resigned to the thought that their honorable work has not captured intelligence, even though the automation of tasks associated with intelligence is convincing.

This realization seems to escape the thinking behind the newest developments in connectionism. The tenor of the day is that what happens in deep and deeper learning is difficult to explain, but is nevertheless intelligent (more or less because we say so). We have voice recognition (of good performance; the margin of error is close to that of human performance in a context of noise), image recognition; we have vision systems of robust performance; there are fraud detection applications working as a matter of routine (in insurance claim evaluations and diagnostic systems). One application leads to another: you identify images either by having them labeled by people playing, over the Internet, some game designed to engage as many as possible in the exercise, or by training networks to recognize them. Others use the knowledge associated with labeling images—e.g., stop sign, work of art, X-ray, scribble—for making new images. The same holds true for sound, video, writing—for anything. It is almost like what the ignition engine made possible: cars: tractors, airplanes, and so much more. But while nobody has claimed that the ignition engine works like the motoric system in the living, almost everyone involved in neural networks posit that they try to emulate how neurons in the brain work when recognizing a face, finding someone in a crowd, hearing a conversation in a noisy room of many conversations, discovering styles of painting or music composition, automatically generating trailers for new movies, etc., etc. Such claims go further: learning of learning, or even a network designing itself for some challenging tasks (mentioned earlier in this text), never mind "neural nets for generating music."[14]

10.2 That's not how the brain works

Recurrent neural networks (RNN), with or without convolutional layers, detect " sentiment" (which for Amazon reviews means positive or negative assessments) the way "a brain would do." Radford, Jozefowicz, and Sutzkever (2017) give away the extent to which such artificial neuron networks use pattern recognition for generative purposes. Images can be generated on the fly; so can sentences be classified (in support of lawyers trying to match arguments to articles of the law). "Remembering" (retrieval, after all) is the desired function of neural networks associated with the so-called "differentiable memory." The Turing machine translated into neural networks (Graves, Wayne, Danihelka 2014) further increases performance for specific applications, as usually explained in relation to how the brain works.

Just for a starter: No, that's not how the brain works. The open-ended variety of neurons, not to say the continuous remaking and the ever-changing map of connections, the integrated nature of all processes—involving the entire body—are only indicative of how the living unfolds. Shallow descriptions of synapses, serving as argument for claiming "The Unreasonable Effectiveness of Recurrent Neural Networks (Karpathy 2015) at best make us aware of the nature of the undecidable nature of cognitive processes. Big Data (from monitoring the body and the brain) and very powerful resources of algorithmic computation are at work in artificial processes to emulate how the human functions, but delivering at most comparable performance devoid of meaning. In the living, the data is scarce. Actually, it is always information, i.e., data associated with meaning, which is totally absent from neural networks (Graves, et al 2016). Living processes are not known for their speed, but rather for a rhythm congruent with life.

---

[14] Kyle McDonald gives many examples. See http://www.kylemcdonald.net/

A good review of deep learning (LeCun, Bengio, Hinton 2015) of less than three years ago is already dated. *arXiv*, a digital platform meant to be a repository, turns descriptions of hundreds of new attempts in deep learning into a cascade of breakthroughs, neither peer-reviewed nor sufficiently clear in their claims. As of the writing of this study, Hinton himself, terribly suspicious of claims of all kind, returned to his capsule networks (Sabour, Frosst, Hinton 2017) as an alternative to the multilayered networks. This in itself is an invitation to a closer look at the entire development in question.

## 11 A Wager in the Age of Deep Learning Euphoria

AlphaGo Zero (Silver et al 2017) gave writers the chance to run away with the trophy: "The AI that has nothing to learn from humans" is the headline (Chan 2017). The memorable "When will AI exceed human performance? Evidence from AI Experts" (Grace et al 2017) opened the fireworks spectacle. Previously, Google's Deep Mind team—the new celebrities of our time of stardom inflation—did not mind writing about *superhuman* performance. In Marvel comics, this makes sense; in Nietsche's *Übermensch*,[15] it marked the death of God—but not a replacement by neural networks. Still, except for a sense of proportion, there is nothing to object over hyperbole reflecting the age in which everybody runs faster and faster after the prize (or is it *price*?). Setting the reporting aside (or simply letting the dust settle), we remain with two open questions:

1. Is reinforcement learning "without human data, guidance or domain knowledge beyond the game rules" really AI?

2. Can this neuronal network that "improves the strength of the tree research" crack the protein folding problem (specifically identified as the next target on which accumulated experience will be used)?

---

[15] Friedrich Nietsche, Thus Spoke Zarathustra: A Book for All and None (Also sprach Zarathustra: Ein Buch für Alle und Keinen) composed in four parts between 1883 and 1885 and published between 1883 and 1891.

## 11.1 The price of mimicking

We have already seen that AI, in its so-called symbolic implementation (propositional knowledge driven) is as artificial as artificial gets, but it is not, even by those who conceived it, intelligence. Between the seductive goal of achieving intelligence in something else than the living and the reality of accomplishments (some impressive beyond what Dreyfus and Weizenbaum, and many others. were willing to acknowledge), there is the realization that automating tasks usually associated with intelligent actions is quite different from engineering intelligence. The automated playing of checkers, with its roughly 500 billion possible positions (Schaeffer et al 2007), or of chess (several times larger space of permutations, which Shannon described quite well) did not involve any intelligence, but rather brute force computation and the appropriate mathematics. That Go, with yet an even larger possible space of choices, posed more challenges is obvious. However, AlphaGo, the winning neuronal net over some of the game's champions, as well as AlphaGo Zero, playing against itself and discovering, unassisted (unsupervised learning) how to play, make us aware of what distinguishes automation from intelligence. One is the domain of data processing. AlphaGo, like Big Blue, picked up patterns from games played previously by masters. AlphaGo Zero was only exposed to what Go is. In the first case, the inductive aspect dominated—not unlike pattern recognition of speech, images, sounds, etc. In the second, deductive and abductive inferences contributed to the success. The program played millions of games against itself. This is how a huge amount of data was generated, and eventually the previous AlphaGo was defeated (100 games to 0).

For the record: the no-data starting point has been attempted before. Indeed, the "self-learning evolutionary chess program (Fogel 2004) and the subsequent steps (2005, 2006) in the direction of a "New Philosophy of Machine Intelligence" (Fogel 1995, 2006) deserve at least some reference. The evaluation function is at the core of evolutionary computation.

The statement that there was nothing to learn from humans ignores that intelligence, informed by cultural interactions, is embodied in the game itself. Its rules encode knowledge. AlphaGo Zero did not invent a new game, and even less a new language of interactions that generalize over cultural or social existence. The novel reinforcement of patterns (which the authors call "learning," in the tradition of machine learning mathematics) algorithm is ingenious, associating probabilities to each move and selecting better choices via a Monte Carlo Tree Search (MCTS). This is reinforcement, but different in nature from that accumulated over the history of people played Go. The narrative—full sequence of moves—is the expression of successful syntax. But when living subjects play the game, they enter the space of story-making, not with the purpose of miming the lottery of large numbers, but rather of reshaping themselves in the experience. With enough computing power, each lottery can be won. Many tried the game when the purse was large enough to justify the expense. But lotteries are only games of chance, not meaningful stories. Each real game tells a new story. There is originality at play, not at willing the lottery, as an expression of the uniqueness of each player. Patterns on the board are sometimes reminiscent of images (one author mentions a structure evoking a lion's mouth or a tortoise shell.) The emphasis is on creativity, not monotonic machine-based success.

11.1.1 The game and the story

The success of Go automation brought back history: the "blood vomiting game" of almost 200 years ago, when the reigning champion—of a Japan different from today's—faced a younger opponent. He lost his life in the extended context of the confrontation. The anecdotal importance is significant for the cultural dimension of any game, from the hide-and-go-seek of our childhood to the new digital games competitions followed by tens of millions. Neural-networks-based applications can beat any computer game, and even the metagames (betting on the outcome or other aspects). This is clear-cut is only proof of their deterministic prowess. Imagine championships of competing neural networks watched by neural networks and adjudicated by yet other networks. Borges must be laughing in his grave: the map literally replaced the territory. Within this world of competing neural networks, there is no need for human experience. But the story is the experience of the journey. Its meaning is the difference between who we were when we first stepped on the path towards the mountain peak, and who we are after we reached it (or even after we gave up trying to reach it). We are our questions. So far, no computer-based application, AI or deep learning, ever formulated a question (never mind a meaningful questions).

Automation of the game produces winning narrations, but never a story. There is no intelligence in the timeline of any event. Time series are not expressions of intelligence, but testimony to action-reaction and the duration—not the time—involved. Stories are expressions of shared intelligence. What succeeds is the meaning, which has no correspondent in any of the sophisticated operations that take place in the neural network.

11.1.2 Target-driven

As a matter of fact, from the simplest to the most complicated network, independent of the techniques used (to which I shall return), we have here an example of a teleological (i.e., driven by the target, the desired outcome) convergence machine. The wedge, the lever, the pulley, and the hammer are as intelligent as any neural network, regardless of the number of layers or of the mathematics (statistics, probability theory, recursion, etc.); likewise, the internal combustion engine. The intelligence is that of the humans who made them, who learned from their use, who perfected them, and who were changed by the experience. None is self-perfecting, from its own resources. None is creative, but can be used creatively. The Otto cycle[16] in the combustion engine is like a neuronal network: it maximizes the net work (no pun intended) that the engine produces.

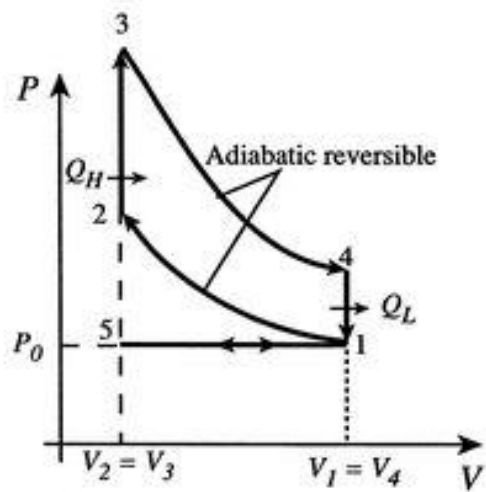

---

[16] http://web.mit.edu/16.unified/www/FALL/thermodynamics/notes/node26.html

**Figure 11** Convergence engineering is of the same nature regardless of the medium for calculation.

This is the target—less complicated than distinguishing from among millions of images, or words, or any other data. The mixture of air and vaporized combustible substance is contained in a cylinder that has a piston at one end. You compress the mixture adiabatically, a spark ignites it, the pressure increases rapidly, the piston is pushed outwards causing expansion, and the work is performed. The equations express the equipartion theorem. The result is clear: an equation describing how the compression ratio (the pressure leading to ignition vs. the pressure of expansion after ignition) affects the efficiency. The engine designers do curve fitting.

Let's summarize what happens in a network (Many details are left out, but not ignored). An artificial neural network (ANN) with hidden layers could be trained in a variety of ways. Imagine building an ignition engine and determining the geometry of the cylinder, the piston, the optimal moment for getting the electrical spark, etc. In the absence of the physics underlying the engine's functioning, you attach weights to your choices: longer, shorter, etc. Actually, an ANN could even be used to optimize the functioning of such an engine. What in deep learning is called *back propagation* is nothing but comparing the desired output (minimum of compression ratio) and the actual output. What propagates is the error—better yet, the lack of knowledge at the beginning, when you seek in the darkness of hunches.

$$\text{error} = \text{function (expected vs. actual output)} \qquad \text{(a)}$$

The input values are weighted (something like *important*, *not so important*, *marginally important*, etc.). Let's say:

$$\text{error} = \text{(desired output – expected output)} \qquad \text{(b)}$$

Given that the expected output is generated under the given selection of a parameter **p** (length of cylinder, or whatever) according to the weight (**w**) attached to it, we end up with

$$\text{error} = (\text{desired output} - \mathbf{w.p}) \qquad\qquad (c)$$

In real life, engineers worked on the engine by adjusting the weight associated with a parameter. In an ANN, the adjustment reflects the mathematics of convergence: A sequence **S$_n$** of numbers (which means data in the interval o to n) is qualified as convergent if it tends to limit S:

$$\lim_{n \to \infty} S_n = S$$

if, for any $\in >0$, there is a number N such that $\vert S_n - S < \Sigma$ for m>N.

There is convergence in any and every algorithmic endeavor; the effective procedure that Hilbert asked about is actually represented by the convergence on the derived result (in his case, the proving of mathematical statements). This can be any repetitive structure, or, as we have already seen, any decidable entity, i.e., that can be fully and consistently described. Feedforward units (frequently convolutional nets) and recurrent networks (with memory components such as LSTM) (Greff et al 2016) execute the mathematics of long sequences of inferences. The number of steps depends on the number of hidden layers in feedforward procedures, and the duration within which a recurrent net recognizes a pattern (process usually described as *remembering*, although no remembering can take place since no membering is possible).

There are also attempts to mimic the living, in the sense that the matter-energy interlocking (mentioned previously in respect to maintenance of unity in diversity) is pursued via energy-base models: attach a score to each possible configuration of the variables. Factor graphs (which are non-probabilistic models) lead to what is called "structured prediction." (An early attempt is Bell

Labs "Graph Transformer Networks," eventually used in reading bank checks and other documents.)

But I do not want to write the history (and pre-history) of deep learning. Much more in this domain is to come. We are prepared to be amazed. What will not change is the reality that machine learning with a well-defined target (as complicated as one chooses) and immense data sets (remember, immense in Elsasser's understanding in describing it) is quite different from how the living operates. The oft-repeated sentence (almost a Credo) is that "We learn from the brain how to get better in deep learning." For the sake of clarity: the theology of the machine is based on circular understanding, and the associated misrepresentations. For anyone with a modicum of knowledge regarding the brain, it is evident that this is not how the brain, or better yet, the integrated organism (of the human being or any other being) works. One (and only one) recent discovery: as the brain and its extension through the entire body is formed, it already gets involved in the making of the organism. Developmental processes are the expression of the aggregate living entity. Birth defects, such as abnormal muscle development (Levin et al. 2017), are an integrated expression of the way all parts of the organism come together. Communication channels from the brain to the body structure are essential to self-repair processes.

11.2 Back to neural networks

The open-ended variety of neurons in the organism is one element. There are no two identical neurons; there are no two identical synapses. Moreover, they are in continuous remaking, some more often, some only once or a few times. The map of their ever-changing map of connections is different in nature from the connections among ANN. The integrated nature of all processes— involving, as just pointed out, the entire organism—stands in no relation to the shallow description in the deep learning maps. A weighted sum of ANN inputs and the involvement of activation function is, of course, a good mathematical tool for describing phenomena of the same nature as those taking place in the artificial network. Error propagation (how one ANN-derived functioning affects the others) is a powerful method that relies on robust mathematics. Calculating the gradient in connection to the network weights and following the gradient in the back propagation is also convincing.

Nevertheless, in the living, targets are continuously changing, and, more important, the data on which the living relies is minimal most of the time. Actually, the living operates on information, i.e., data associated with meaning. The rather high expense of energy to achieve what the living performs naturally, with limited resources, is indicative of the illusions of deep learning as the new frontier of AI.

Various neural network aficionados have taken note of the fact that AlphaGo (in the Fan and Lee configurations) were distributed over 176 GPUs and 48 TPUs, respectively. AlphaGo Zero and AlphaGo Master run on a single machine with 4 TPU. TPU stands for the Tensor Processing Unit developed by Google. One remark: "It took about 30+ days of wallclock to train. That's about 110 megawatt hours (MWh) worth of energy required."[17] That translates into over 500 years of a person learning how to play Go. Having been affirmed that the energy balance reflects the law of conservation, it is not surprising that what took centuries to become a culturally shared language (that of the game) takes a huge amount of energy to be mimicked!

---

[17] Hacker News: Google supercharges machine learning tasks with TPU custom chip (googleblg.com), https://news.ycombinator.com/item?id=11724763. See also: The relationship between clockspeed and power consumption is nonlinear, https://electronics.stackexchange.com/questions/122050/what-limits-cpu-speed

11.3 A Wager

AlphaGo Zero, or whatever follows along the line of deterministic algorithmic computation (whether in deep learning configuration or any other form), might break many records, or might open all kinds of new avenues. Some mention sparse coding: a kind of standard technique similar to dynamic programming. Others have their eyes on proof procedures, in the sense that interconnections (characteristic of complicated phenomena conjuring complicated mathematical descriptions or others) and networking seem congenial.

The major players in the world economy (Google, IBM, Apple, Intel, GM, Samsung, Nvidia, etc.) have acquired deep learning start-ups, often without really knowing why. Those active in the field are not disillusioned characters. From academia or from research facilities, they exercise a great deal of influence (including access to public and private funding), but also warn about misunderstandings. François Chollet put it quite bluntly: "Current supervised perception and reinforcement learning algorithms require lots of data, are terrible at planning, and are only doing straightforward pattern recognition (Yao 2017). The contrast is to the human being, able to "learn from very few examples, can do very long-term planning, and are capable of forming abstract models of a situation…." Spoiled by the qualifier "father of deep learning" (I'm sure he would give credit at least to Rumelhart), Hinton points to evolution in describing "features that are early in a sensory pathway"[18]—an idea that Bernstein documented decades back, in the early stages of anticipation (Nadin 2015), and which Fogel applied (as mentioned above). Against this background of realistic self-assessment appear various so-called contributions that run against the elementary understanding of how ANN and their variations—probabilistic PNN, time-delay TDNN, convolutional CNN (even better, ConvNet), recurrent RNN, Hopfield network, and Boltzmann machine, self-organizing map SOM, long short-term memory LSTM, and others— actually work. The initial claims, suggestive of what happens in the living, deserve at most an ironic "of course," since biological connections and mathematically inspired artifacts share more in the label than in anything else.

---

[18] See http://opsbug.com/deep-learning

The syntactic condition (which undermined symbolic AI) is, by the nature of the processes triggered, impossible to transcend. There is no semantic (at most the pseudo-semantics delivered by ontology engineers), and there is no pragmatics. Does it really matter if such a device understands why a vehicle has to stop at the red traffic light?

11.3.1 Understanding is not optional

The so-called *idiot-savant* (a term redefined as *autistic savant* in order to avoid offense) is like some of the most impressive networks. For example, they "memorize telephone directories, they know exactly how many matches fell from the box, etc., etc. These are precise operations performed without any of idea of meaning, and even less of how ("I saw how many cards ere in the…"). Every amazing performance of an *idiot-savant* that we are aware of can be accomplished in the ANN domain. But there is no intelligence to it. Rather, once we analyze it, after the event, we find that lack of intelligence characterizes the unusual (even phenomenal) performance. In reference to a totally different domain: you can get the most attractive reproductions of impressionist paintings (and other artworks) in Dafen, China, where the largest mass producer—60% of the global volume—of knock-off canvases is located. There is no difference between the syntactic level at which copies of paintings by van Gogh, Monet, Picasso, etc. are produced and the manner in which neural networks do the same. Do they understand what they do, in the manner in which those who created the copied images understood them? Is there any creativity to be identified in the effort?

This question can be repeated *ad nauseum* in respect to many other activities often labeled "intelligent." To qualify as intelligent, an action has to be performed within an understanding of what it means. Otherwise, it is an automated procedure for what intelligent entities would perform—or maybe not—given a context as large as the culture, social norms, political values, etc. in which it operates. When Lin, Tegmark, and Rolnick (2017) ask "Why does deep and cheap learning work so well?" they correctly point to its mathematics and physics. The probability that an image (of which each pixel can take 256 values) is a cat relates to the $256^{1,000,000}$ probabilities (an immense number, larger than that of atoms in the universe, ca. $10^{78}$). As the authors describe the process, the ANN "performs a combinatorial swindle": exponentiation (at the power of one million!) is replaced by multiplication. The 256 values that each pixel can take no longer leads to $\mathbf{v}^n$ (exponentiation), but to $\mathbf{v}$ x $\mathbf{n}$ (multiplication). As already mentioned in the previous section, the living works with rather few values, but they are significant. In the network, there is no way to assess significance. An infant recognizes a cat in the pragmatic space, not by searching endlessly in the mathematical domain of syntactic possibilities. Neural networks are closed to meaning.

With this in mind, the rather audacious target of addressing (point 2 in the prior section) based on the impressive Learn-Go-By-Yourself deep learning implementation appears questionable under the light of adequacy, more than under the light of performance—no matter how energy intensive.

11.4 Decidable vs. undecidable

**Thesis**: If artificial neural networks, in whichever configuration, could be used to explain (never mind anticipate) protein folding, so could any deterministic device, such as the ignition engine or the hammer (both already mentioned).

Of course, this is a statement open to many interpretations, including the understanding that neural networks, as deterministic instantiations of data processing are not capable of describing protein folding as an anticipation driven process (no repetition of any kind!).

It is at this juncture that the G-complexity perspective I advanced (Nadin 2013, 2014, 2017a) again begs for attention. Gödel's fundamental distinction between the decidable and the undecidable generalized to reality provides the criterion for defining the living, embodied in matter, in contradistinction to the non-living. Physics provides effective tools and methods for nomothetically describing the non-living. This includes, of course, cause-and-effect causality, eventually challenged within a quantum mechanic's view. Distinctions grounded in the realization of stochastic processes allowed for refinements that will continue to be made. The non-living can be fully and consistently described. Maybe the qualifiers "fully" and even "consistently" are a bit too sharp, but it is beyond question that the non-living is decidable—even if at quantum level, or at large scale interstellar dynamics, more distinctions would prove necessary.

The living is undecidable. Causality in the living does not exclude the deterministic aspect of the non-living, but rather complements it. Where the non-living is the realm of action-reaction, the living is defined by anticipation: action driven by a possible future (sometimes called *retrocausation*, Werbos 2008). The reductionist-deterministic view of non-living matter was confirmed, over centuries of experiment: either as thought-experiments (*Gedankenexperimente*) or in the setting of measurements that translate the decidable into numbers subject to confirmation through replication.

To gain knowledge of the living—which means of ourselves as we change through each of our actions, including measurement and experiment—is by the nature of life not subject to the same method that applies to the non-living. This is why replicating experiments concerning the dynamics of life—a very valuable gnoseological activity—is to add observation to the narrative of the living subject to such experiments. (For more details on this issue, see Nadin 2017b.) Feynman, probably the most active missionary of a physics-based view of everything, realized that computation—i.e., automated mathematics—is not congenial to every form of knowledge it delivers. He argued in favor of embodying computation in the reality we want to simulate. Most well known was his call to facilitate knowledge of quantum mechanics by embodying computation in quantum phenomena (the so-called quantum computer would be the actual embodiment) (Feynman, 1982).

Protein folding is among the living processes through which life is defined. The unfolding of the stem cell is another example. They are anticipatory in the sense that no folding is a repeat, as no act of creation via the stem cell results in any identical outcome. This is yet another instantiation, empirically proven, for even a longer time than any experiment pertinent to the non-living, of "repetition without repetition" characteristic of the undecidable. G-complexity defines life. A repeat statement: Anticipation is couched in G-complexity. The threshold—the decidable vs. the undecidable—is, like everything else in reality, probably less clear-cut than we might wish (used as we are by now to the universe of zero and one, or Yes and No). But it is effective to the extent that computation embodied in the non-living is decidable—and this includes the infinite loop problem. Computation in the living is undecidable.

That IARPA, in its misleading call for research of what they define as Anticipatory Intelligence does not understand the meaning of anticipation is probably amusing. I answered my colleague, who congratulated me for IARPA's catching up: "Thank heaven, they are not. The distance is increasing. Why would I wish to have the intelligence community trailing me?"—well, they do, but that is a different story. That AlphaGo Zero even considers unlocking the secrets of protein folding is, however, epistemologically aggravating. It is like proving that Earth is flat against all evidence to the contrary. Their convergence machine is pretty good. I feel happy to be proven right that neural networks are convergence machines. From among all those practitioners in the field with whom I have shared my definition, two felt that there was something to it. Hinton wrote to me, "Neural nets converge to a state that locally minimizes the error." Patrick Eklund also wrote (September 23, 2017) "The sequence of parameter values converge through iteration. Optimization is a convergence (but convergence is not always optimization)." Well, *protein folding is the opposite of convergence; it is infinite divergence, there is no repetition of any fold.*

## 12 Beyond the Oxymoron

Dreyfus and Weizenbaum tried hard to point to the epistemological aspects of intelligence: we gain access to knowledge, but as we get knowledge, we change. Our own change testifies to intelligence, not as a prerequisite, but rather as an outcome. They did not want to undermine the courageous attempt to define AI, but rather they were not yet convinced that the task of automation is the same as intelligence. They failed—although the public echo was pretty telling of how the subject of algorithmic computation was perceived. Dreyfus was honored many times, and became a member of the American Academy of Sciences. Weizenbaum, refusing to enlist his talents in projects for the military establishment, became a hero in exile. But before them, America ignored Charles Sanders Peirce—just as they ignored, while they lived, Robert Rosen and even, more recently, Lotfi Zadeh, whose ideas[19] were more successful abroad than in his adopted country.

Those who can afford it —Wolfram, with his 2,3 Turing machine, Loebner, and others--have challenged computational science with all kinds of prizes. Turing himself, not necessarily accepted for what he was, initiated a test for acknowledging intelligence. My own wager is not one of reputation, but of epistemological coherence. It is based on the following: access to knowledge of life and the living means access to the representation of the undecidable and the understanding of life as an expression sui-generis of creativity. Means of expression of a decidable nature can only extract the decidable component, thereby falsifying the knowledge. As a scientist, I refuse myself the luxury of stating what seems (at some moment in time) impossible.

Deep learning, in whichever implementation (they are getting more and more sophisticated) could open access to decidable aspects of it, but not to the undecidable process based upon which self-preservation of life as a creative process is manifested. This is a matter of perspective, which I wish my talented distinguished colleagues in deep learning would consider.

---

[19] Including the early formulation: Thinking Machines – A new field in electrical engineering, Columbia Engineering Quarterly, January 1950.

The fundamental characteristic of life is its continuous remaking, re-creation within its self-preservation dynamics. Therefore, claims regarding creativity as an output of neural networks are not to be taken lightly. Deep learning automated pattern recognition through convergence at a high cost of energy and other resources. To claim automation of creativity—"We created new forms of art"—comes close to "We created life." This goes beyond the oxymoron: "We made god and god made us," based upon which the theology of the machine is (still!) practiced.

**Acknowledgments instead of any conclusion**

This study was in progress when in the summer of 2017 Karamjit Gill announced a memorial issue dedicated to Huber Dreyfus's legacy. It is the outcome of almost 30 years of work in computation—writing programs, testing ideas, carrying out experiments—and of no less intense dedication to understanding how computation has changed us. During this long preparation, I experienced Dreyfus's prosopagnosia three times. Indeed, he could not recognize me (as he had the same problems with others). My enthusiasm for computation made him often lose patience. He wanted to write a review of *The Civilization of Illiteracy*, but in the end could not find time for it. Weizenbaum imparted to me many insights into academic life: you can have a chair at MIT, but if you do not bring in the money, there was no electricity in the room where the chair was located. In Hamburg (Mediale, 1998) and later in Berlin (2004), we disagreed as only Talmudic scholars would—mainly because Weizenbaum and I were into debunking the rapidly growing mythology of the "mother of all machines." Some of the thoughts in my text go back to conversations with both.

Other conversations—with McCarthy and Minsky—a short exchange with Simon, and another with Pat Hayes are also reflected in the text. Over many years, Lotfi Zadeh listened patiently to my arguments and shared some of his own with me, challenging me with his examples. Also over many years, Pamela McCorduck and Terry Winograd assisted, not always agreeing with what I had to say.

No, this study was not supported by any grant, except that of my wife's generous willingness to the sounding board for ideas that would not qualify as middle of the road statements—and sometimes coming up with her own. I remain responsible for all my inferences, faulty or otherwise.